\documentclass[11pt, a4paper, logo, copyright, nonumbering]{deepseek}
\usepackage[authoryear, sort&compress, round]{natbib}
\usepackage{dblfloatfix}
\usepackage{ulem}
\usepackage{caption}
\usepackage{dramatist}
\usepackage{xspace}
\usepackage{pifont}
\usepackage{multirow}
\usepackage{tcolorbox}
\usepackage{xltabular}
\usepackage{longtable}
\usepackage{hyperref}
\usepackage{amsfonts}
\usepackage{amsmath}
\usepackage{amssymb}
\usepackage{lineno}

\usepackage[bottom]{footmisc}

\usepackage{CJKutf8}
\usepackage{subfigure}
\usepackage{setspace}

\usepackage{dsfont}

\newcommand{\spmoe}{DeepSeekMoE}
\newcommand{\tabincell}[2]{\begin{tabular}{@{}#1@{}}#2\end{tabular}}

\makeatletter
\def\@BTrule[#1]{%
  \ifx\longtable\undefined
    \let\@BTswitch\@BTnormal
  \else\ifx\hline\LT@hline
    \nobreak
    \let\@BTswitch\@BLTrule
  \else
     \let\@BTswitch\@BTnormal
  \fi\fi
  \global\@thisrulewidth=#1\relax
  \ifnum\@thisruleclass=\tw@\vskip\@aboverulesep\else
  \ifnum\@lastruleclass=\z@\vskip\@aboverulesep\else
  \ifnum\@lastruleclass=\@ne\vskip\doublerulesep\fi\fi\fi
  \@BTswitch}
\makeatother

\addto\extrasenglish{
}

 {\begin{list}{}%
         {\setlength{\leftmargin}{#1}}%
         \item[]%
 }
 {\end{list}}
 
\reportnumber{001} 

\renewcommand{\today}{}

\title{
\spmoe{}: Towards Ultimate Expert Specialization in 
\\
~~~~~~~~~~~~~~~~~~~~~Mixture-of-Experts Language Models
}

\author[*]{
\small
Damai Dai$^{*1,2}$,
Chengqi Deng$^{1}$,
Chenggang Zhao$^{*1,3}$,
R.X. Xu$^{1}$,
Huazuo Gao$^{1}$,
Deli Chen$^{1}$, 
Jiashi Li$^{1}$,
Wangding Zeng$^{1}$,
Xingkai Yu$^{*1,4}$,
Y. Wu$^{1}$,
Zhenda Xie$^{1}$,
Y.K. Li$^{1}$,
Panpan Huang$^{1}$,
Fuli Luo$^{1}$,
Chong Ruan$^{1}$,
\newline
Zhifang Sui$^{2}$,
Wenfeng Liang$^{1}$
\\
\small
$^1$DeepSeek-AI \\
\small
$^2$National Key Laboratory for Multimedia Information Processing, Peking University \\
\small
$^3$Institute for Interdisciplinary Information Sciences, Tsinghua University \\
\small
$^4$National Key Laboratory for Novel Software Technology, Nanjing University \\
\small
\texttt{\{daidamai, szf\}@pku.edu.cn},
\small
\texttt{\{wenfeng.liang\}@deepseek.com}
\\
\small
\url{https://github.com/deepseek-ai/DeepSeek-MoE}
}
\correspondingauthor{Contribution during internship at DeepSeek-AI.}

\renewcommand{\phi}{\varphi}

\renewcommand{\leq}{\leqslant}

\renewcommand{\epsilon}{\varepsilon}
\renewcommand{\imath}{\mathrm{i}}

\newlength{\restsubwidth}
\newlength{\restsubheight}
\newlength{\restsubmoreheight}
\newcommand{\rest}[2]{%
        \settowidth{\restsubwidth}{\ensuremath{#2}}
        \settoheight{\restsubheight}{\ensuremath{{}_{#2}}}
        \ensuremath{{#1\hskip 0.5pt}_{\vrule\kern2pt\parbox[b][%
        4pt][b]{\the\restsubwidth}{%
                        \ensuremath{{}_{#2}}}}}
        }

\begin{abstract}

In the era of large language models, Mixture-of-Experts (MoE) is a promising architecture for managing computational costs when scaling up model parameters. 
However, conventional MoE architectures like GShard, which activate the top-$K$ out of $N$ experts, face challenges in ensuring expert specialization, i.e. each expert acquires non-overlapping and focused knowledge.
In response, we propose the \textbf{\spmoe{}} architecture towards ultimate expert specialization. 
It involves two principal strategies: 
(1) finely segmenting the experts into $mN$ ones and activating $mK$ from them, allowing for a more flexible combination of activated experts; 
(2) isolating $K_s$ experts as shared ones, aiming at capturing common knowledge and mitigating redundancy in routed experts.
Starting from a modest scale with 2B parameters, we demonstrate that \spmoe{} 2B achieves comparable performance with GShard 2.9B, which has 1.5$\times$ expert parameters and computation. 
In addition, \spmoe{} 2B nearly approaches the performance of its dense counterpart with the same number of total parameters, which set the upper bound of MoE models. 
Subsequently, we scale up \spmoe{} to 16B parameters and show that it achieves comparable performance with LLaMA2 7B, with only about 40\% of computations. 
Further, our preliminary efforts to scale up \spmoe{} to 145B parameters consistently validate its substantial advantages over the GShard architecture, and show its performance comparable with DeepSeek 67B, using only 28.5\% (maybe even 18.2\%) of computations.

\end{abstract}

\begin{document}
\begin{CJK*}{UTF8}{gbsn}

\maketitle

\begin{figure}[!ht]
\centering
\includegraphics[width=0.7\linewidth]{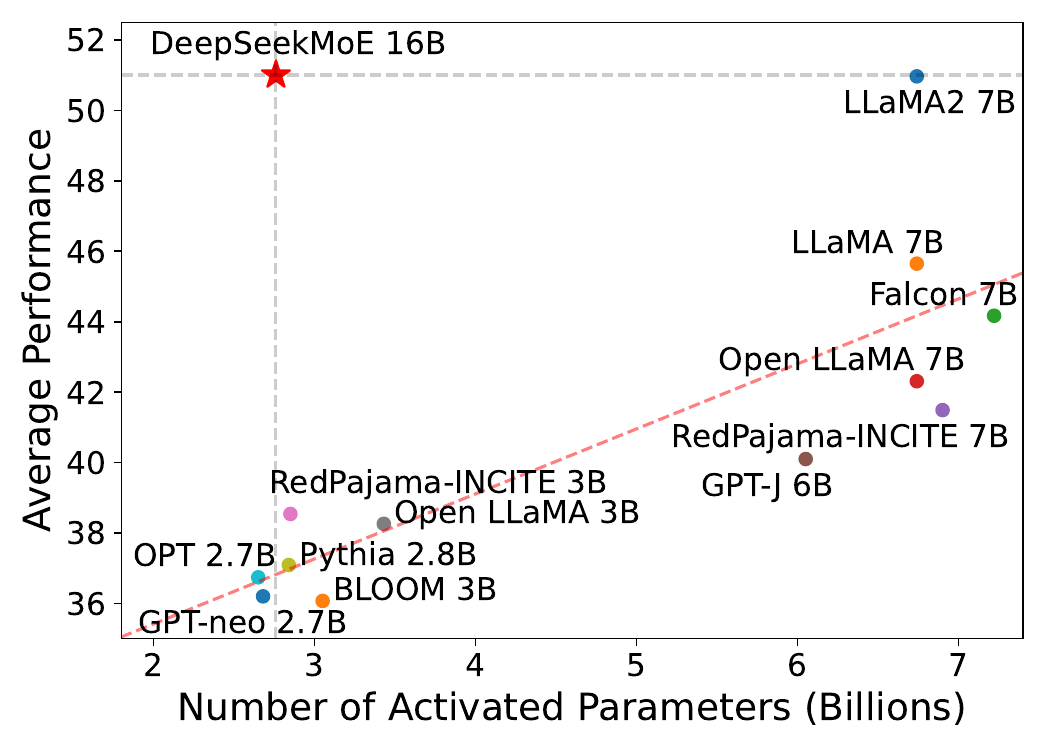}
\caption{
Comparison between \spmoe{} 16B and open source models on the Open LLM Leaderboard. 
The red dashed line is linearly fitted from data points of all models except \spmoe{} 16B.
\spmoe{} 16B consistently outperforms models with a similar number of activated parameters by a large margin, and achieves comparable performance with LLaMA2 7B, which has approximately 2.5 times the activated parameters.
}
\label{fig:openllm}
\end{figure}

\section{Introduction}

Recent research and practices have empirically demonstrated that, with sufficient training data available, scaling language models with increased parameters and computational budgets can yield remarkably stronger models~\citep{gpt3,gpt4,llama,scaling_law}. 
It is imperative to acknowledge, however, that the endeavor to scale models to an extremely large scale is also associated with exceedingly high computational costs.
Considering the substantial costs, the Mixture-of-Experts~(MoE) architecture~\citep{ori_moe1,ori_moe2,moe} has emerged as a popular solution.
It can enable parameter scaling, while concurrently keeping computational costs at a modest level.
Recent applications of MoE architectures in Transformers~\citep{transformer} have yielded successful attempts at scaling language models to a substantial size~\citep{switch,gshard,glam,st_moe}, accompanied with remarkable performance. 
These achievements underscore the considerable potential and promise of MoE language models.

Despite the promising potential of MoE architectures, existing MoE architectures potentially suffer from issues of knowledge hybridity and knowledge redundancy, which limit the expert specialization, i.e., each expert acquires non-overlapping and focused knowledge.
Conventional MoE architectures substitute the Feed-Forward Networks (FFNs) in a Transformer with MoE layers. 
Each MoE layer consists of multiple experts, with each structurally identical to a standard FFN, and each token is assigned to one~\citep{switch} or two~\citep{gshard} experts. 
This architecture manifests two potential issues:
(1) 
\textbf{Knowledge Hybridity}: existing MoE practices often employ a limited number of experts (e.g., 8 or 16), and thus tokens assigned to a specific expert will be likely to cover diverse knowledge. 
Consequently, the designated expert will intend to assemble vastly different types of knowledge in its parameters, which are hard to utilize simultaneously.
(2) 
\textbf{Knowledge Redundancy}: tokens assigned to different experts may require common knowledge. 
As a result, multiple experts may converge in acquiring shared knowledge in their respective parameters, thereby leading to redundancy in expert parameters. 
These issues collectively hinder the expert specialization in existing MoE practices, preventing them from reaching the theoretical upper-bound performance of MoE models.

In response to the aforementioned issues, we introduce \textbf{\spmoe{}}, an innovative MoE architecture specifically designed towards ultimate expert specialization. 
Our architecture involves two principal strategies: 
(1) \textbf{Fine-Grained Expert Segmentation:} 
while maintaining the number of parameters constant, we segment the experts into a finer grain by splitting the FFN intermediate hidden dimension. 
Correspondingly, keeping a constant computational cost, we also activate more fine-grained experts to enable a more flexible and adaptable combination of activated experts.
Fine-grained expert segmentation allows diverse knowledge to be decomposed more finely and be learned more precisely into different experts, where each expert will retain a higher level of specialization. 
In addition, the increased flexibility in combining activated experts also contributes to a more accurate and targeted knowledge acquisition.
(2) \textbf{Shared Expert Isolation:}
we isolate certain experts to serve as shared experts that are always activated, aiming at capturing and consolidating common knowledge across varying contexts. 
Through compressing common knowledge into these shared experts, redundancy among other routed experts will be mitigated. 
This can enhance the parameter efficiency and ensure that each routed expert retains specialized by focusing on distinctive aspects.
These architectural innovations in \spmoe{} offer opportunities to train a parameter-efficient MoE language model where each expert is highly specialized. 

Starting from a modest scale with 2B parameters, we validate the advantages of the \spmoe{} architecture. 
We conduct evaluations on 12 zero-shot or few-shot benchmarks spanning diverse tasks. 
Empirical results indicate that \spmoe{} 2B surpasses GShard 2B~\citep{gshard} by a substantial margin, and even matches GShard 2.9B, a larger MoE model with 1.5$\times$ expert parameters and computation. 
Remarkably, we find that \spmoe{} 2B nearly approaches the performance of its dense counterpart with an equivalent number of parameters, which sets the strict upper bound of MoE language models.
In pursuit of deeper insights, we conduct elaborate ablation studies and analysis on the expert specialization for \spmoe{}. 
These studies validate the effectiveness of fine-grained expert segmentation and shared expert isolation, and provide empirical evidence supporting the assertion that \spmoe{} can achieve a high level of expert specialization.  

Leveraging our architecture, we subsequently scale up the model parameters to 16B and train \spmoe{} 16B on a large-scale corpus with 2T tokens. 
Evaluation results reveal that with only about 40\% of computations, \spmoe{} 16B achieves comparable performance with DeepSeek 7B~\citep{deepseek_llm}, a dense model trained on the same 2T corpus. 
We also compare \spmoe{} with open source models and the evaluations demonstrate that \spmoe{} 16B consistently outperforms models with a similar number of activated parameters by a large margin, and achieves comparable performance with LLaMA2 7B~\citep{llama2}, which has approximately 2.5 times the activated parameters. 
Figure~\ref{fig:openllm} demonstrates the evaluation results on the Open LLM Leaderboard\footnote{https://huggingface.co/spaces/HuggingFaceH4/open\_llm\_leaderboard}. 
Additionally, we conduct supervised fine-tuning~(SFT) for alignment, transforming the model into a chat model.
Evaluation results show that \spmoe{} Chat 16B also achieves comparable performance with DeepSeek Chat 7B and LLaMA2 SFT 7B in the chat setting. 
Encouraged by these results, we further undertake a preliminary endeavor to scale up \spmoe{} to 145B. 
The experimental results still validate its substantial advantages over the GShard architecture consistently.
In addition, it shows performance comparable with DeepSeek 67B, using only 28.5\% (maybe even 18.2\%) of computations.

Our contributions are summarized as follows: 
\begin{itemize}
    \item \textbf{Architectural Innovation.}
    We introduce \spmoe{}, an innovative MoE architecture aiming at achieving ultimate expert specialization, which employs two principal strategies of fine-grained expert segmentation and shared expert isolation. 
    \item \textbf{Empirical Validation.}
    We conduct extensive experiments to empirically validate the effectiveness of the \spmoe{} architecture. 
    Experimental results validate the high level of expert specialization in \spmoe{} 2B, and indicate that \spmoe{} 2B can nearly approach the upper bound performance for MoE models
    \item \textbf{Scalability.}
    We scale up \spmoe{} to train a 16B model and show that with only about 40\% of computations, \spmoe{} 16B achieves comparable performance with DeepSeek 7B and LLaMA2 7B. 
    We also undertake a preliminary endeavor to scale up \spmoe{} to 145B, highlighting its consistent advantages over the GShard architecture and showing a comparable performance with DeepSeek 67B.
    \item \textbf{Alignment for MoE.}
    We successfully perform supervised fine-tuning on \spmoe{} 16B to create an aligned chat model, showcasing the adaptability and versatility of \spmoe{} 16B.
    \item \textbf{Public Release.}
    In the spirit of open research, we release the model checkpoint of \spmoe{} 16B to the public. 
    Notably, this model can be deployed on a single GPU with 40GB of memory without the need for quantization. 
\end{itemize}

\section{Preliminaries: Mixture-of-Experts for Transformers}
\label{sec:preliminary}

We first introduce a generic MoE architecture commonly used in Transformer language models. 
A standard Transformer language model is constructed by stacking $L$ layers of standard Transformer blocks, where each block can be represented as follows:
\begin{align}
    \mathbf{u}_{1:T}^{l} &= \operatorname{Self-Att}\left( \mathbf{h}_{1:T}^{l-1} \right) + \mathbf{h}_{1:T}^{l-1}, \\
    \mathbf{h}_{t}^{l} &= \operatorname{FFN}\left( \mathbf{u}_{t}^{l} \right) + \mathbf{u}_{t}^{l},
\end{align}
where $T$ denotes the sequence length, 
$\operatorname{Self-Att}(\cdot)$ denotes the self-attention module, 
$\operatorname{FFN}(\cdot)$ denotes the Feed-Forward Network~(FFN), 
$\mathbf{u}_{1:T}^{l} \in \mathbb{R}^{T \times d}$ are the hidden states of all tokens after the $l$-th attention module, 
and $\mathbf{h}_{t}^{l} \in \mathbb{R}^{d}$ is the output hidden state of the $t$-th token after the $l$-th Transformer block. 
For brevity, we omit the layer normalization in the above formulations. 

A typical practice to construct an MoE language model usually substitutes FFNs in a Transformer with MoE layers at specified intervals~\citep{switch,gshard,glam,st_moe}.
An MoE layer is composed of multiple experts, where each expert is structurally identical to a standard FFN.
Then, each token will be assigned to one~\citep{switch} or two~\citep{gshard} experts. 
If the $l$-th FFN is substituted with an MoE layer, the computation for its output hidden state $\mathbf{h}_{t}^{l}$ is expressed as:
\begin{align}
\mathbf{h}_{t}^{l} & = \sum_{i=1}^{N} \left( {g_{i,t} \operatorname{FFN}_{i}\left( \mathbf{u}_{t}^{l} \right)} \right) + \mathbf{u}_{t}^{l}, \\
g_{i,t} & = \begin{cases} 
s_{i,t}, & s_{i,t} \in \operatorname{Topk} (\{ s_{j, t} | 1 \leq j \leq N \}, K), \\
0, & \text{otherwise}, 
\end{cases} \\
s_{i,t} & = \operatorname{Softmax}_i \left( {\mathbf{u}_{t}^{l}}^{T} \mathbf{e}_{i}^{l} \right), 
\end{align}
where $N$ denotes the total number of experts, 
$\operatorname{FFN}_{i}(\cdot)$ is the $i$-th expert FFN, 
$g_{i,t}$ denotes the gate value for the $i$-th expert, 
$s_{i,t}$ denotes the token-to-expert affinity, 
$\operatorname{Topk}(\cdot, K)$ denotes the set comprising $K$ highest affinity scores among those calculated for the $t$-th token and all $N$ experts,
and $\mathbf{e}_{i}^{l}$ is the centroid of the $i$-th expert in the $l$-th layer. 
Note that $g_{i,t}$ is sparse, indicating that only $K$ out of $N$ gate values are nonzero. 
This sparsity property ensures computational efficiency within an MoE layer, i.e., each token will be assigned to and computed in only $K$ experts.
Also, in the above formulations, we omit the layer normalization operation for brevity. 

\begin{figure}[ht]
\centering
\includegraphics[width=0.99\linewidth]{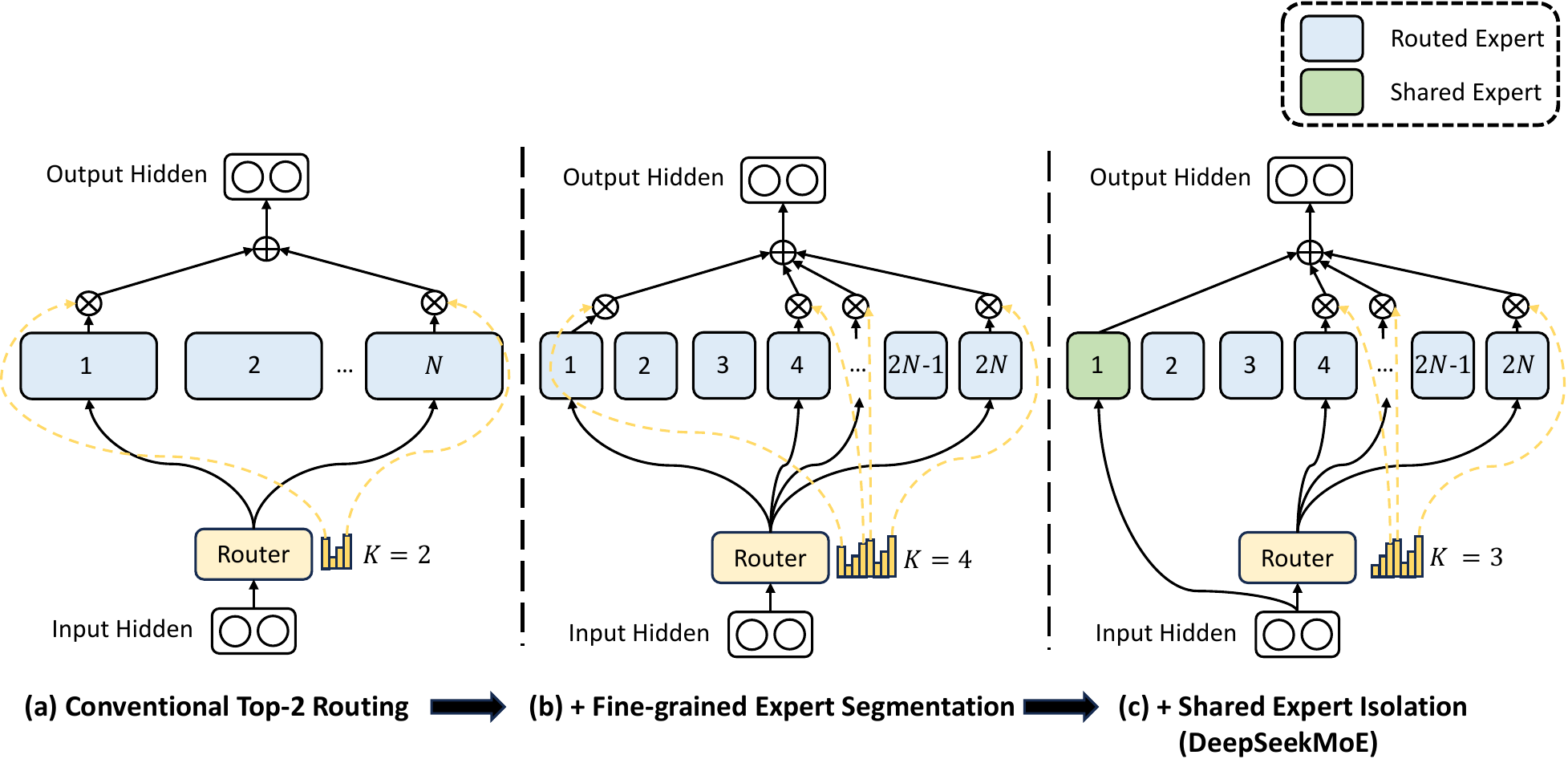}
\caption{
Illustration of \spmoe{}. 
Subfigure (a) showcases an MoE layer with the conventional top-2 routing strategy. 
Subfigure (b) illustrates the fine-grained expert segmentation strategy. 
Subsequently, subfigure (c) demonstrates the integration of the shared expert isolation strategy, constituting the complete \spmoe{} architecture.
It is noteworthy that across these three architectures, the number of expert parameters and computational costs remain constant. 
}
\label{fig:deepseek_moe}
\end{figure}

\section{\spmoe{} Architecture}

On top of the generic MoE architecture outlined in Section~\ref{sec:preliminary}, we introduce \spmoe{}, which is specifically designed to exploit the potential of expert specialization. 
As illustrated in Figure~\ref{fig:deepseek_moe}, our architecture incorporates two principal strategies: fine-grained expert segmentation and shared expert isolation. 
Both of these strategies are designed to elevate the level of expert specialization.

\subsection{Fine-Grained Expert Segmentation}

In scenarios where the number of experts is limited, tokens assigned to a particular expert will be more likely to cover diverse types of knowledge. 
As a consequence, the designated expert will intend to learn vastly different types of knowledge in its parameters, and they are hard to be simultaneously utilized.
However, if each token can be routed to more experts, diverse knowledge will gain the potential to be decomposed and learned in different experts respectively. 
In this context, each expert can still retain a high level of expert specialization, contributing to a more focused knowledge distribution across experts.

In pursuit of the goal, while maintaining a consistent number of expert parameters and computational cost, we segment the experts with a finer grain. 
The finer expert segmentation enables a more flexible and adaptable combination of activated experts. 
To be specific, on top of a typical MoE architecture shown in Figure~\ref{fig:deepseek_moe}(a), we segment each expert FFN into $m$ smaller experts by reducing the FFN intermediate hidden dimension to $\frac{1}{m}$ times its original size. 
Since each expert becomes smaller, in response, we also increase the number of activated experts to $m$ times to keep the same computation cost, as illustrated in Figure~\ref{fig:deepseek_moe}(b).
With the fine-grained expert segmentation, the output of an MoE layer can be expressed as:
\begin{align}
\mathbf{h}_{t}^{l} & = \sum_{i=1}^{mN} \left( {g_{i,t} \operatorname{FFN}_{i}\left( \mathbf{u}_{t}^{l} \right)} \right) + \mathbf{u}_{t}^{l}, \\
g_{i,t} & = \begin{cases} 
s_{i,t}, & s_{i,t} \in \operatorname{Topk} (\{ s_{j, t} | 1 \leq j \leq mN \}, mK), \\
0, & \text{otherwise}, 
\end{cases} \\
s_{i,t} & = \operatorname{Softmax}_i \left( {\mathbf{u}_{t}^{l}}^{T} \mathbf{e}_{i}^{l} \right), 
\end{align}
where the total number of expert parameters is equal to $N$ times the number of parameters in a standard FFN, and $mN$ denotes the total number of fine-grained experts. 
With the fine-grained expert segmentation strategy, the number of nonzero gates will also increases to $mK$. 

From a combinatorial perspective, the fine-grained expert segmentation strategy substantially enhances the combinatorial flexibility of activated experts.
As an illustrative example, we consider the case where $N=16$. 
A typical top-2 routing strategy can yield $\binom{16}{2}=120$ possible combinations.
By contrast, if each expert is split into $4$ smaller experts, the fine-grained routing strategy can yield $\binom{64}{8}=4,426,165,368$ potential combinations. 
The surge in combinatorial flexibility enhances the potential for achieving more accurate and targeted knowledge acquisition. 

\subsection{Shared Expert Isolation}

With a conventional routing strategy, tokens assigned to different experts may necessitate some common knowledge or information. 
As a result, multiple experts may converge in acquiring shared knowledge in their respective parameters, thereby resulting in redundancy in expert parameters. 
However, if there are shared experts dedicated to capturing and consolidating common knowledge across varying contexts, the parameter redundancy among other routed experts will be alleviated. 
This alleviation of redundancy will contribute to a more parameter-efficient model with more specialized experts.

Towards this objective, in addition to the fine-grained expert segmentation strategy, we further isolate $K_{s}$ experts to serve as shared experts.
Regardless of the router module, each token will be deterministically assigned to these shared experts.
In order to maintain a constant computational cost, the number of activated experts among the other routed experts will be decreased by $K_{s}$, as depicted in Figure~\ref{fig:deepseek_moe}(c).
With the shared expert isolation strategy integrated, an MoE layer in the complete \spmoe{} architecture is formulated as follows:
\begin{align}
\mathbf{h}_{t}^{l} & = \sum_{i=1}^{K_{s}} {\operatorname{FFN}_{i}\left( \mathbf{u}_{t}^{l} \right)} + \sum_{i=K_{s} + 1}^{mN} \left( {g_{i,t} \operatorname{FFN}_{i}\left( \mathbf{u}_{t}^{l} \right)} \right) + \mathbf{u}_{t}^{l}, \\
g_{i,t} & = \begin{cases} 
s_{i,t}, & s_{i,t} \in \operatorname{Topk} (\{ s_{j, t} | K_{s} + 1 \leq j \leq mN \}, mK - K_{s}), \\
0, & \text{otherwise}, 
\end{cases} \\
s_{i,t} & = \operatorname{Softmax}_i \left( {\mathbf{u}_{t}^{l}}^{T} \mathbf{e}_{i}^{l} \right). 
\end{align}
Finally, in \spmoe{}, the number of shared expert is $K_{s}$, 
the total number of routed experts is $mN - K_{s}$, 
and the number of nonzero gates is $mK - K_{s}$. 

It is worth noting that the prototype of shared expert isolation can be credited to \citet{deepspeed_moe}. 
The key distinction lies in the fact that they derive this strategy from an engineering perspective, while we approach it from an algorithmic standpoint. 

\subsection{Load Balance Consideration}

Automatically learned routing strategies may encounter the issue of load imbalance, which manifests two notable defects. 
Firstly, there is a risk of routing collapse~\citep{moe}, i.e.,  the model always selects only a few experts, preventing other experts from sufficient training. 
Secondly, if experts are distributed across multiple devices, load imbalance can exacerbate computation bottlenecks.

\paragraph{Expert-Level Balance Loss.}
In order to mitigate the risk of routing collapse, we also employ an expert-level balance loss. 
The computation of the balance loss is as follows:
\begin{align}
    \mathcal{L}_{\mathrm{ExpBal}} & = \alpha_1 \sum_{i=1}^{N^{\prime}}{f_i P_i}, \\
    f_i & = \frac{N^{\prime}}{K^{\prime}T} \sum_{t=1}^{T}{ \mathds{1}( \text{Token $t$ selects Expert $i$} )}, \\
    P_i & = \frac{1}{T} \sum_{t=1}^{T}{s_{i,t}},
\end{align}
where $\alpha_1$ is a hyper-parameter called expert-level balance factor, 
$N^{\prime}$ is equal to $(mN - K_s)$ and $K^{\prime}$ is equal to $(mK - K_s)$ for brevity. 
$\mathds{1}(\cdot)$ denotes the indicator function. 

\paragraph{Device-Level Balance Loss.}
In addition to the expert-level balance loss, we introduce a device-level balance loss. 
When aiming to alleviate computation bottlenecks, it becomes unnecessary to enforce strict balance constraints at the expert level, because excessive constraints on load balance will compromise model performance. 
Instead, our primary objective is to ensure balanced computation across the devices.
If we partition all routed experts into $D$ groups $\{\mathcal{E}_1, \mathcal{E}_2, ..., \mathcal{E}_D \}$, and deploy each group on a single device, the device-level balance loss is computed as follows:
\begin{align}
    \mathcal{L}_{\mathrm{DevBal}} & = \alpha_{2} \sum_{i=1}^{D}{f_i^{\prime} P_i^{\prime}}, \\
    f_i^{\prime} & = \frac{1}{|\mathcal{E}_i|} \sum_{j \in \mathcal{E}_i}{ f_j }, \\
    P_i^{\prime} & = \sum_{j \in \mathcal{E}_i}{ P_j },
\end{align}
where $\alpha_{2}$ is a hyper-parameter called device-level balance factor. 
In practice, we set a small expert-level balance factor to mitigate the risk of routing collapse, and meanwhile set a larger device-level balance factor to promote balanced computation across the devices.

\section{Validation Experiments}

\subsection{Experimental Setup}

\subsubsection{Training Data and Tokenization}
\label{sec:training_data}

Our training data is sampled from a large-scale multilingual corpus created by DeepSeek-AI. 
The corpus primarily focuses on English and Chinese but also encompasses other languages. 
It is derived from diverse sources, including web text, mathematical material, coding scripts, published literature, and various other textual materials. 
For the purpose of validation experiments, we sample a subset containing 100B tokens from the corpus to train our models.
For tokenization, we utilize the HuggingFace Tokenizer\footnote{https://github.com/huggingface/tokenizers} tools to train byte pair encoding (BPE)~\citep{bpe} tokenizers on a smaller subset of the training corpus. 
In the validation experiments, we prepare a tokenizer with a vocabulary size of 8K, and the vocabulary size will be scaled up when training larger models.

\subsubsection{Infrastructures}

We conduct experiments based on HAI-LLM~\citep{haillm}, an efficient and light-weight training framework which integrates multiple parallelism strategies, including tensor parallelism~\citep{megatron, megatron2, megatron3}, ZeRO data parallelism~\citep{zero}, PipeDream pipeline parallelism~\citep{pipedream}, and more specifically, expert parallelism~\citep{gshard} by combining data and tensor parallelism.
In order to optimize performance, we develop GPU kernels with CUDA and Triton~\citep{triton} for gating algorithms and fusing computations across linear layers in different experts.

All experiments are carried out on clusters equipped with NVIDIA A100 or H800 GPUs.
Each node in the A100 cluster contains 8 GPUs connected pairwise via the NVLink bridge. 
The H800 cluster also features 8 GPUs per node, interconnected using NVLink and NVSwitch within nodes. 
For both A100 and H800 clusters, InfiniBand interconnects are utilized to facilitate communication across nodes.

\subsubsection{Hyper-Parameters}

\paragraph{Model Settings.}
In the validation experiments, we set the number of Transformer layers to 9 and the hidden dimension to 1280. 
We employ the multi-head attention mechanism with a total of 10 attention heads, where each head has a dimension of 128. 
For initialization, all learnable parameters are randomly initialized with a standard deviation of 0.006.
We substitute all FFNs with MoE layers, and ensure that the total number of expert parameters equals 16 times that of a standard FFN. 
Additionally, we keep the activated expert parameters, including shared expert parameters and activated routed expert parameters, as 2 times that of a standard FFN.
Under this configuration, each MoE model has approximately 2B total parameters, with the number of activated parameters around 0.3B.

\paragraph{Training Settings.}
We employ the AdamW optimizer~\citep{adamw} with hyper-parameters set to $\beta_1=0.9$, $\beta_2=0.95$, and $\mathrm{weight\_decay}=0.1$. 
The learning rate is scheduled using a warmup-and-step-decay strategy. 
Initially, the learning rate linearly increases from 0 to the maximum value during the first 2K steps. 
Subsequently, the learning rate is multiplied by 0.316 at 80\% of the training steps, and again by 0.316 at 90\% of the training steps. 
The maximum learning rate for validation experiments is set to $1.08 \times 10^{-3}$, and the gradient clipping norm is set to 1.0.
The batch size is set to 2K, and with a maximum sequence length of 2K, each training batch contains 4M tokens. 
Correspondingly, the total number of training steps is set to 25,000 to achieve 100B training tokens. 
Due to the abundance of training data, we do not use dropout during training. 
Given the relatively small model size, all parameters, including expert parameters, are deployed on a single GPU device to avoid unbalanced computation. 
Correspondingly, we do not drop any tokens during training and do not employ the device-level balance loss. 
In order to prevent routing collapse, we set an expert-level balance factor of 0.01.

For readability, we also present an overview table of hyper-parameters for \spmoe{} across different sizes in Appendix~\ref{sec:app_hyper_params}.

\subsubsection{Evaluation Benchmarks}

We conduct evaluations on a wide range of benchmarks covering various types of tasks. 
We list the benchmarks as follows. 

\paragraph{Language Modeling.}
For language modeling, we evaluate the models on the test set of Pile~\citep{pile}, and the evaluation metric is the cross-entropy loss. 

\paragraph{Language Understanding and Reasoning.}
For language understanding and reasoning, we consider HellaSwag~\citep{hellaswag}, PIQA~\citep{piqa}, ARC-challenge and ARC-easy~\citep{arc}. 
The evaluation metric for these tasks is accuracy.

\paragraph{Reading Comprehension.} 
For reading comprehension, we use RACE-high and RACE-middle \cite{race}, and the evaluation metric is accuracy. 

\paragraph{Code Generation.} 
For code generation, we evaluate the models on HumanEval~\citep{codex} and MBPP~\citep{mbpp}. 
The evaluation metric is Pass@1, which represents the pass rate for only one generation attempt.

\paragraph{Closed-Book Question Answering.} 
For closed-book question answering, we consider TriviaQA~\citep{triviaqa} and NaturalQuestions~\citep{nq}. 
The evaluation metric is the Exactly Matching~(EM) rate. 

\begin{table}[ht]
\centering
\setlength{\tabcolsep}{5pt}
\begin{tabular}{@{}l c| c c c | c c@{}}
\toprule
\textbf{Metric} & \textbf{\# Shot} & \textbf{Dense} & \textbf{Hash Layer} & \textbf{Switch} & \textbf{GShard} & \textbf{\spmoe{}} \\
\midrule
\# Total Params & N/A & 0.2B & 2.0B & 2.0B & 2.0B & 2.0B \\
\# Activated Params & N/A & 0.2B & 0.2B & 0.2B & 0.3B & 0.3B \\
FLOPs per 2K Tokens & N/A & 2.9T & 2.9T & 2.9T & 4.3T & 4.3T \\
\# Training Tokens & N/A & 100B & 100B & 100B & 100B & 100B \\
\midrule
Pile~(Loss) & N/A & 2.060 & 1.932 & 1.881 & 1.867 & \textbf{1.808} \\
\midrule
HellaSwag~(Acc.) & 0-shot & 38.8 & 46.2 & 49.1 & 50.5 & \textbf{54.8} \\
PIQA~(Acc.) & 0-shot & 66.8 & 68.4 & 70.5 & 70.6 & \textbf{72.3} \\
ARC-easy~(Acc.) & 0-shot & 41.0 & 45.3 & 45.9 & 43.9 & \textbf{49.4} \\
ARC-challenge~(Acc.) & 0-shot & 26.0 & 28.2 & 30.2 & 31.6 & \textbf{34.3} \\
\midrule
RACE-middle~(Acc.) & 5-shot & 38.8 & 38.8 & 43.6 & 42.1 & \textbf{44.0} \\
RACE-high~(Acc.) & 5-shot & 29.0 & 30.0 & 30.9 & 30.4 & \textbf{31.7} \\
\midrule
HumanEval~(Pass@1) & 0-shot & 0.0 & 1.2 & 2.4 & 3.7 & \textbf{4.9} \\
MBPP~(Pass@1) & 3-shot & 0.2 & 0.6 & 0.4 & 0.2 & \textbf{2.2} \\
\midrule
TriviaQA~(EM) & 5-shot & 4.9 & 6.5 & 8.9 & 10.2 & \textbf{16.6} \\
NaturalQuestions~(EM) & 5-shot & 1.4 & 1.4 & 2.5 & 3.2 & \textbf{5.7} \\
\bottomrule
\end{tabular}
\caption{
Evaluation results for validation experiments. 
\textbf{Bold} font indicates the best. 
Compared with other MoE architectures, \spmoe{} exhibits a substantial performance advantage.
}
\label{tab:valid_exp_main}
\end{table}

\subsection{Evaluations}

\paragraph{Baselines.}
Including \spmoe{}, we compare five models for validation experiments. 
\textbf{Dense} denotes a standard dense Transformer language model with 0.2B total parameters. 
\textbf{Hash Layer}~\citep{hash} is an MoE architecture based on top-1 hash routing, with 2.0B total parameters and 0.2B activated parameters, aligned with the dense baseline. 
\textbf{Switch Transformer}~\citep{switch} is another well-known MoE architecture based on top-1 learnable routing, with total parameters and activated parameters the same as Hash Layer. 
\textbf{GShard}~\citep{gshard} employs a top-2 learnable routing strategy, with 2.0B total parameters and 0.3B activated parameters since one more expert is activated compared to top-1 routing methods. 
\textbf{\spmoe{}} has 1 shared expert and 63 routed experts, where each expert is 0.25 times the size of a standard FFN. 
Including \spmoe{}, all compared models share the same training corpus and training hyper-parameters. 
All compared MoE models have the same number of total parameters, and GShard has the same number of activated parameters as \spmoe{}.

\paragraph{Results.}
We present the evaluation results in Table~\ref{tab:valid_exp_main}.
For all demonstrated models, we report the final evaluation results after training on 100B tokens. 
From the table, we make the following observations:
(1) With sparse architectures and more total parameters, Hash Layer and Switch Transformer achieve significantly stronger performance than the dense baseline with the same number of activated parameters.
(2) Compared with Hash Layer and Switch Transformer, GShard has more activated parameters and achieves slightly better performance than Switch Transformer.
(3) With the same number of total parameters and activated parameters, \spmoe{} demonstrates overwhelming advantages over GShard. 
These results showcase the superiority of our \spmoe{} architecture within the existing landscape of MoE architectures.

\begin{table}[ht]
\centering
\setlength{\tabcolsep}{8pt}
\begin{tabular}{@{}l c | c c c c@{}}
\toprule
\textbf{Metric} & \textbf{\# Shot} & \textbf{GShard$\times 1.5$} & \textbf{Dense$\times 16$} & \textbf{\spmoe{}} \\
\midrule
Relative Expert Size & N/A & 1.5 & 1 & 0.25 \\
\# Experts & N/A & 0 + 16 & 16 + 0 & 1 + 63 \\
\# Activated Experts & N/A & 0 + 2 & 16 + 0 & 1 + 7 \\
\# Total Expert Params & N/A & 2.83B & 1.89B & 1.89B \\
\# Activated Expert Params & N/A & 0.35B & 1.89B & 0.24B \\
FLOPs per 2K Tokens & N/A & 5.8T & 24.6T & 4.3T \\
\# Training Tokens & N/A & 100B & 100B & 100B \\
\midrule
Pile~(Loss) & N/A & 1.808 & 1.806 & 1.808 \\
\midrule
HellaSwag~(Acc.) & 0-shot & 54.4 & 55.1 & 54.8 \\
PIQA~(Acc.) & 0-shot & 71.1 & 71.9 & 72.3 \\
ARC-easy~(Acc.) & 0-shot & 47.3 & 51.9 & 49.4 \\
ARC-challenge~(Acc.) & 0-shot & 34.1 & 33.8 & 34.3 \\
\midrule
RACE-middle~(Acc.) & 5-shot & 46.4 & 46.3 & 44.0 \\
RACE-high~(Acc.) & 5-shot & 32.4 & 33.0 & 31.7 \\
\midrule
HumanEval~(Pass@1) & 0-shot & 3.0 & 4.3 & 4.9 \\
MBPP~(Pass@1) & 3-shot & 2.6 & 2.2 & 2.2 \\
\midrule
TriviaQA~(EM) & 5-shot & 15.7 & 16.5 & 16.6 \\
NaturalQuestions~(EM) & 5-shot & 4.7 & 6.3 & 5.7 \\
\bottomrule
\end{tabular}
\caption{
Comparisons among \spmoe{}, larger GShard models, and larger dense models.
In the line of ``\# Experts'', $a$ + $b$ denotes $a$ shared experts and $b$ routed experts. 
In the line of ``\# Activated Experts'', $a$ + $b$ denotes $a$ activated shared experts and $b$ activated routed experts. 
\spmoe{} achieves comparable performance with a GShard model containing 1.5 times expert parameters and computation.
In addition, \spmoe{} nearly approaches the performance of a dense model with 16 times FFN parameters, which sets the upper bound for MoE models in terms of the model capacity. 
}
\label{tab:valid_exp_larger}
\end{table}

\subsection{\spmoe{} Aligns Closely with the upper bound of MoE Models}

We have demonstrated that \spmoe{} outperforms the dense baseline and other MoE architectures. 
In order to provide a more precise understanding of the performance of \spmoe{}, we compare it with larger baselines with more total parameters or activated parameters. 
The comparisons enable us to estimate the required model size of GShard or dense baselines to achieve equivalent performance to \spmoe{}.

\paragraph{Comparison with GShard$\times 1.5$.}
Table~\ref{tab:valid_exp_larger} shows the comparison between \spmoe{} and a larger GShard model with 1.5 times the expert size, which results in 1.5 times both expert parameters and expert computation. 
Overall, we observe that \spmoe{} achieves comparable performance with GShard$\times 1.5$, underscoring the significant advantage inherent in the \spmoe{} architecture.
In addition to the comparison with GShard$\times 1.5$, we also show the comparison with GShard$\times 1.2$ in Appendix~\ref{sec:app_compare_larger}. 

Furthermore, we increase the number of total parameters of \spmoe{} to 13.3B and compare it with GShard$\times 1.2$ and GShard$\times 1.5$ with 15.9B and 19.8B total parameters, respectively. 
We find that at a larger scale, \spmoe{} can even outperform GShard$\times 1.5$ distinctly. 
These results are also provided in Appendix~\ref{sec:app_compare_larger}. 

\paragraph{Comparison with Dense$\times 16$.}
Table~\ref{tab:valid_exp_larger} also shows the comparison between \spmoe{} and larger dense models. 
For a fair comparison, we do not use the widely used ratio (1:2) between the attention and FFN parameters. 
Instead, we configure 16 shared experts where each expert has the same number of parameters as a standard FFN. 
This architecture mimics a dense model with 16 times standard FFN parameters. 
From the table, we find that \spmoe{} nearly approaches the performance of Dense$\times 16$, which sets the strict upper bound of MoE models in terms of the model capacity. 
These results suggest that, \textbf{\textit{at least at the scale of about 2B parameters and 100B training tokens, the performance of \spmoe{} aligns closely with the theoretical upper bound of MoE models}}. 
Also, we provide additional comparisons with Dense$\times 4$ in Appendix~\ref{sec:app_compare_larger}.

\begin{figure}[ht]
\centering
\includegraphics[width=0.99\linewidth]{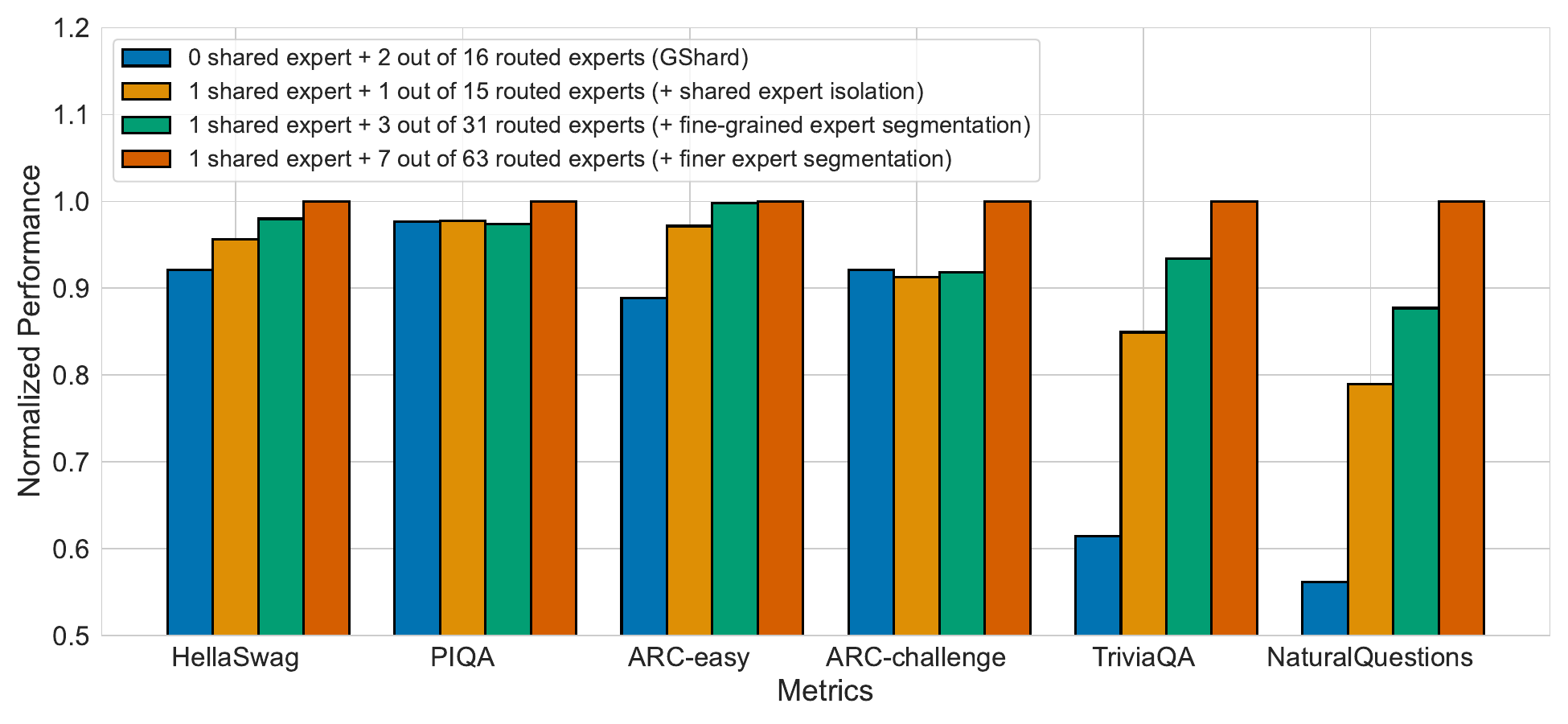}
\caption{
Ablation studies for \spmoe{}. 
The performance is normalized by the best performance for clarity in presentation. 
All compared models have the same number of parameters and activated parameters. 
We can find that fine-grained expert segmentation and shared expert isolation both contribute to stronger overall performance. 
}
\label{fig:valid_exp_ablation}
\end{figure}

\subsection{Ablation Studies}

In order to substantiate the effectiveness of the fine-grained expert segmentation and shared expert isolation strategies, we conduct ablation studies for \spmoe{} and present the results in Figure~\ref{fig:valid_exp_ablation}. 
For a fair comparison, we ensure all models included in the comparison have the same number of total parameters and activated parameters.

\paragraph{Shared Expert Isolation.}
In order to evaluate the influence of the shared expert isolation strategy, we isolate one expert as the shared one based on GShard. 
From Figure~\ref{fig:valid_exp_ablation}, we observe that compared with GShard, the intentional isolation of a shared expert yields improved performance across a majority of benchmarks. 
These results support the proposition that the shared expert isolation strategy contributes to a stronger model performance. 

\paragraph{Fine-Grained Expert Segmentation.}
In order to assess the effectiveness of the fine-grained expert segmentation strategy, we conduct a more detailed comparison by further segmenting the experts into a finer grain. 
To be specific, we segment each expert into 2 or 4 smaller experts, resulting in a total of 32 (1 shared + 31 routed) or 64 (1 shared + 63 routed) experts. 
Figure~\ref{fig:valid_exp_ablation} reveals a consistent trend that the continuous refinement of expert segmentation granularity corresponds to a continuous enhancement in overall model performance. 
These findings provide empirical substantiation for the effectiveness of the fine-grained expert segmentation strategy. 

\paragraph{Ratios Between Shared and Routed Experts.}
In addition, we investigate the best ratio of shared experts and routed experts. 
Based on the finest granularity with 64 total experts and keeping the number of total experts and activated experts constant, we attempt to isolate 1, 2, and 4 experts as shared ones. 
We find that different ratios of the shared experts and routed experts do not significantly impact the performance, and 1, 2, and 4 shared experts achieve a Pile loss of 1.808, 1.806, and 1.811, respectively. 
Considering that the ratio of 1:3 yields a marginally better Pile loss, when scaling up \spmoe{}, we keep the ratio between shared experts and activated routed experts as 1:3. 

\subsection{Analysis on Expert Specialization}
\label{sec:validation_analysis_specialization}

In this section, we conduct an empirical analysis on the expert specialization of \spmoe{} 2B. 
\spmoe{} 2B in this section refers to the model reported in Table~\ref{tab:valid_exp_main}, i.e., comprising 2.0B total parameters, with 1 shared expert and 7 out of 63 routed experts being activated. 

\begin{figure}[ht]
\centering
\includegraphics[width=0.6\linewidth]{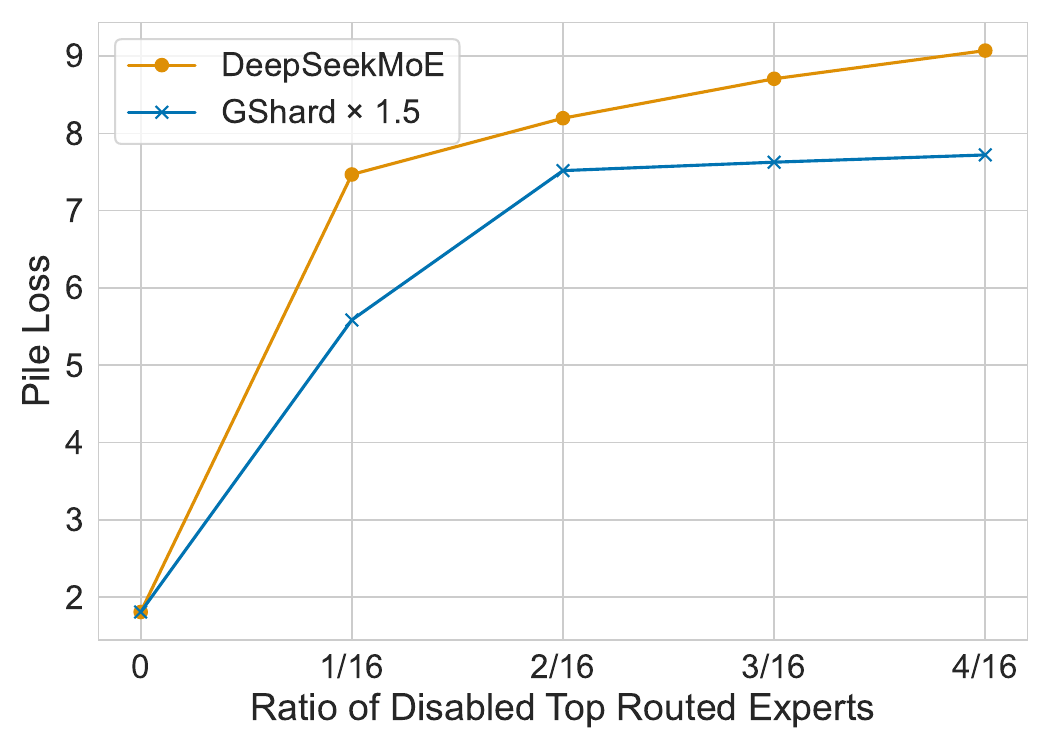}
\caption{
Pile loss with regard to different ratios of disabled top routed experts. 
Notably, \spmoe{} exhibits greater sensitivity to the ratio of disabled top routed experts, indicating lower redundancy among routed experts in \spmoe{}.
}
\label{fig:valid_redundancy}
\end{figure}

\paragraph{\spmoe{} Exhibits Lower Redundancy Among Routed Experts.}
In order to assess the redundancy among routed experts, we disable varying ratios of top routed experts and evaluate the Pile loss. 
To be specific, for each token, we mask a certain ratio of experts with the highest routing probability, and then select top-K experts from the remaining routed experts.
For fairness, we compare \spmoe{} with GShard$\times 1.5$ since they have the same Pile loss when no experts are disabled. 
As shown in Figure~\ref{fig:valid_redundancy}, compared with GShard$\times 1.5$, \spmoe{} is more sensitive to the disabling of top routed experts. 
This sensitivity suggests a lower level of parameter redundancy in \spmoe{}, since each routed expert is more irreplaceable. 
In contrast, GShard$\times 1.5$ exhibits greater redundancy among its expert parameters, so it can buffer the performance drop when top routed experts are disabled.

\paragraph{Shared Experts Are Irreplaceable by Routed Experts.}
In order to investigate the role of the shared expert in \spmoe{}, we disable it and activate one more routed expert. 
The evaluation on Pile shows a significant increase in the Pile loss, rising from 1.808 to 2.414, even though we maintain the same computational cost. 
This result highlights the crucial function of the shared expert and indicates that the shared expert captures fundamental and essential knowledge not shared with routed experts, making it irreplaceable by routed ones. 

\begin{figure}[ht]
\centering
\includegraphics[width=0.6\linewidth]{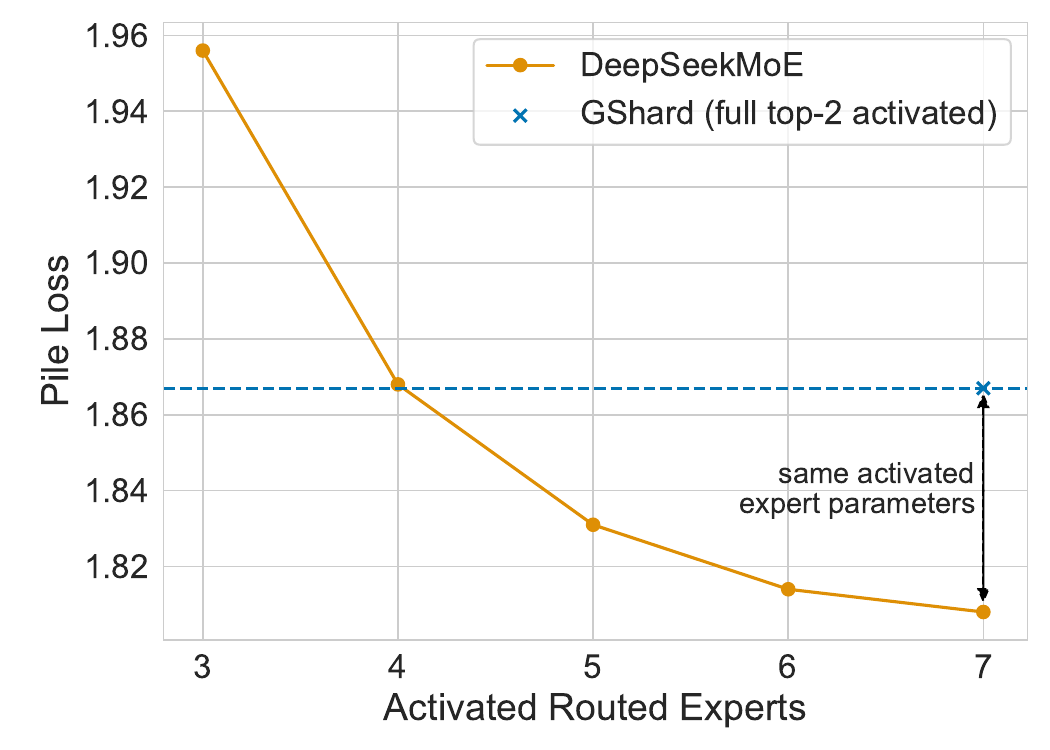}
\caption{
Pile loss with regard to different numbers of activated routed experts in \spmoe{}. 
With only 4 routed experts activated, \spmoe{} achieves a Pile loss comparable with GShard. 
}
\label{fig:valid_partial_activated}
\end{figure}

\begin{figure}[!ht]
\centering
\includegraphics[width=0.99\linewidth]{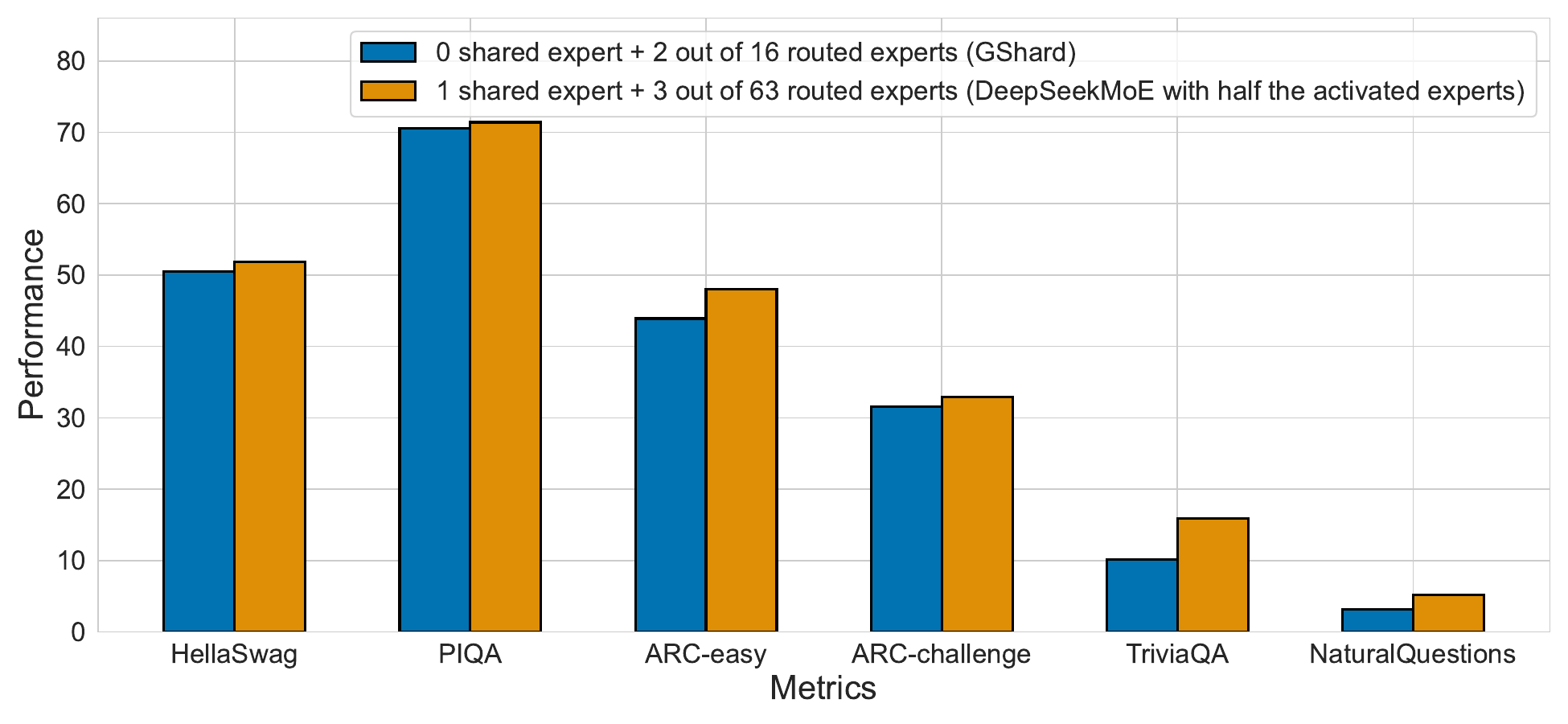}
\caption{
Comparison between GShard and \spmoe{} with half the activated experts (trained from scratch). 
With the same total expert parameters and only half of the activated expert parameters, \spmoe{} still outperforms GShard. 
}
\label{fig:valid_less_activated}
\end{figure}

\paragraph{\spmoe{} Acquires Knowledge More Accurately.}
In order to validate our claim that higher flexibility in combining activated experts contributes to a more accurate and targeted knowledge acquisition, we investigate whether \spmoe{} can acquire requisite knowledge with fewer activated experts.
To be specific, we vary the number of activated routed experts from 3 to 7 and evaluate the resulting Pile loss. 
As demonstrated in Figure~\ref{fig:valid_partial_activated}, even with only 4 routed experts activated, \spmoe{} achieves a Pile loss comparable with GShard. 
This observation supports the proposition that \spmoe{} can acquire requisite knowledge more accurately and efficiently.

Encouraged by these findings, in order to validate the expert specialization and accurate knowledge acquisition of \spmoe{} more rigorously, we train a new model from scratch. 
This model comprises 1 shared expert and 63 routed experts, where only 3 routed experts are activated. 
The evaluation results shown in Figure~\ref{fig:valid_less_activated} demonstrate that, even with the same total expert parameters and only half of the activated expert parameters, \spmoe{} still outperforms GShard. 
This highlights the ability of \spmoe{} to leverage expert parameters more efficiently, i.e., the proportion of effective parameters in the activated experts is much higher than that of GShard. 

\section{Scaling up to \spmoe{} 16B}

With the \spmoe{} architecture, we scale up our MoE model to a larger scale with 16B total parameters and train it on 2T tokens. 
Our results demonstrate that compared with LLaMA2 7B, \spmoe{} 16B achieves superior performance with only about 40\% of computations.

\subsection{Experimental Setup}

\subsubsection{Training Data and Tokenization}

We sample the training data from the same corpus as described in Section~\ref{sec:training_data}. 
Different from the validation experiments, we sample a larger amount of data with 2T tokens, aligning with the number of training tokens of LLaMA2 7B.  
We also use the HuggingFace Tokenizer tools to train a BPE tokenizer, but the vocabulary size is set to 100K for \spmoe{} 16B. 

\subsubsection{Hyper-Parameters}

\paragraph{Model Settings.}
For \spmoe{} 16B, we set the number of Transformer layers to 28 and the hidden dimension to 2048. 
We employ the multi-head attention mechanism with a total of 16 attention heads, where each head has a dimension of 128. 
As for initialization, all learnable parameters are randomly initialized with a standard deviation of 0.006.
We substitute all FFNs except for the first layer with MoE layers, since we observe that the load balance status converges especially slower for the first layer. 
Each MoE layer consists of 2 shared experts and 64 routed experts, where each expert is 0.25 times the size of a standard FFN. 
Each token will be routed to these 2 shared experts and 6 out of 64 routed experts. 
An even finer expert segmentation granularity is not employed due to the potential reduction in computational efficiency associated with excessively small expert sizes. 
At a larger scale over 16B, a finer granularity can still be employed.
Under our configuration, \spmoe{} 16B has approximately 16.4B total parameters, with the number of activated parameters around 2.8B.

\paragraph{Training Settings.}
We employ the AdamW optimizer~\citep{adamw} with hyper-parameters set to $\beta_1=0.9$, $\beta_2=0.95$, and $\mathrm{weight\_decay}=0.1$. 
The learning rate is also scheduled using a warmup-and-step-decay strategy. 
Initially, the learning rate linearly increases from 0 to the maximum value during the first 2K steps. 
Subsequently, the learning rate is multiplied by 0.316 at 80\% of the training steps, and again by 0.316 at 90\% of the training steps. 
The maximum learning rate for \spmoe{} 16B is set to $4.2 \times 10^{-4}$, and the gradient clipping norm is set to 1.0.
The batch size is set to 4.5K, and with a maximum sequence length of 4K, each training batch contains 18M tokens. 
Correspondingly, the total number of training steps is set to 106,449 to achieve 2T training tokens. 
Due to the abundance of training data, we do not use dropout during training. 
We leverage pipeline parallelism to deploy different layers of a model on different devices, and for each layer, all the experts will be deployed on the same device. 
Therefore, we also do not drop any tokens during training and do not employ the device-level balance loss. 
In order to prevent routing collapse, we set a quite small expert-level balance factor of 0.001 because we find that under our parallelization strategy, a higher expert-level balance factor cannot increase the computation efficiency, but instead, it will compromise the model performance. 

\subsubsection{Evaluation Benchmarks}
\label{sec:16b_benchmark}

In addition to the benchmarks used in the validation experiments, we incorporate additional benchmarks for a more comprehensive evaluation. 
We introduce the distinctions from the benchmarks used in validation experiments as follows. 

\paragraph{Language Modeling.}
For language modeling, we also evaluate the models on the test set of Pile~\citep{pile}. 
Since the tokenizer used in \spmoe{} 16B is different from that used in LLaMA2 7B. 
For a fair comparison, we use bits per byte (BPB) as the evaluation metric. 

\paragraph{Reading Comprehension.} 
For reading comprehension, we additionally consider DROP~\citep{drop}. 
The evaluation metric is the Exactly Matching~(EM) rate. 

\paragraph{Math Reasoning.} 
For math reasoning, we additionally incorporate GSM8K~\citep{gsm8k} and MATH~\citep{math}, using EM as the evaluation metric. 

\paragraph{Multi-Subject Multiple-Choice.}
For multi-subject multiple-choice, we additionally evaluate the models on MMLU~\citep{mmlu}. 
The evaluation metric is accuracy. 

\paragraph{Disambiguation.} 
For disambiguation, we additionally consider WinoGrande~\citep{winogrande} and the evaluation metric is accuracy. 

\paragraph{Chinese Benchmarks.}
Since \spmoe{} 16B is pretrained on a bilingual corpus, we also evaluate it on four Chinese benchmarks. 
CLUEWSC~\citep{clue} is a Chinese disambiguation benchmark. 
CEval~\citep{ceval} and CMMLU~\citep{cmmlu} are two Chinese multi-subject multiple-choice benchmarks with a similar form to MMLU. 
CHID~\citep{chid} is a Chinese idiom completion benchmark, aiming to evaluate the understanding of Chinese culture.
The evaluation metrics for the aforementioned Chinese benchmarks are accuracy or EM. 

\paragraph{Open LLM Leaderboard.}
We evaluate all of the aforementioned benchmarks based on our internal evaluation framework. 
In order to compare \spmoe{} 16B with open source models fairly and conveniently, we additionally evaluate \spmoe{} 16B on the Open LLM Leaderboard. 
The Open LLM Leaderboard is a public leaderboard supported by HuggingFace, it consists of six tasks: ARC~\citep{arc}, HellaSwag~\citep{hellaswag}, MMLU~\citep{mmlu}, TruthfulQA~\citep{truthfulqa}, Winogrande~\citep{winogrande}, and GSM8K~\citep{gsm8k}.

\subsection{Evaluations}

\begin{table}[ht]
\centering
\setlength{\tabcolsep}{4pt}
\begin{tabular}{@{}l c| c c@{}}
\toprule
\textbf{Metric} & \textbf{\# Shot} & \textbf{DeepSeek 7B (Dense)} & \textbf{\spmoe{} 16B} \\
\midrule
\# Total Params & N/A & 6.9B & 16.4B \\
\# Activated Params & N/A & 6.9B & 2.8B \\
FLOPs per 4K Tokens & N/A & 183.5T & 74.4T \\
\# Training Tokens & N/A & 2T & 2T \\
\midrule
Pile~(BPB) & N/A & 0.75 & \textbf{0.74} \\
\midrule
HellaSwag~(Acc.) & 0-shot & 75.4 & \textbf{77.1} \\
PIQA~(Acc.) & 0-shot & 79.2 & \textbf{80.2} \\
ARC-easy~(Acc.) & 0-shot & \textbf{67.9} & \textbf{68.1} \\
ARC-challenge~(Acc.) & 0-shot & 48.1 & \textbf{49.8} \\
\midrule
RACE-middle~(Acc.) & 5-shot & \textbf{63.2} & 61.9 \\
RACE-high~(Acc.) & 5-shot & \textbf{46.5} & \textbf{46.4} \\
DROP~(EM) & 1-shot & \textbf{34.9} & 32.9 \\
\midrule
GSM8K~(EM) & 8-shot & 17.4 & \textbf{18.8} \\
MATH~(EM) & 4-shot & 3.3 & \textbf{4.3} \\
\midrule
HumanEval~(Pass@1) & 0-shot & 26.2 & \textbf{26.8} \\
MBPP~(Pass@1) & 3-shot & \textbf{39.0} & \textbf{39.2} \\
\midrule
TriviaQA~(EM) & 5-shot & 59.7 & \textbf{64.8} \\
NaturalQuestions~(EM) & 5-shot & 22.2 & \textbf{25.5} \\
\midrule
MMLU~(Acc.) & 5-shot & \textbf{48.2} & 45.0 \\
\midrule
WinoGrande~(Acc.) & 0-shot & \textbf{70.5} & \textbf{70.2} \\
\midrule
\midrule
CLUEWSC~(EM) & 5-shot & \textbf{73.1} & 72.1 \\
CEval~(Acc.) & 5-shot & \textbf{45.0} & 40.6 \\
CMMLU~(Acc.) & 5-shot & \textbf{47.2} & 42.5 \\
CHID~(Acc.) & 0-shot & \textbf{89.3} & \textbf{89.4} \\
\bottomrule
\end{tabular}
\caption{
Comparison between DeepSeek 7B and \spmoe{} 16B. 
\textbf{Bold} font indicates the best or near the best.
With only 40.5\% of computations, \spmoe{} 16B achieves comparable performance with DeepSeek 7B. 
}
\label{tab:16b_exp_deepseek7b}
\end{table}

\subsubsection{Internal Comparison with DeepSeek 7B}
\label{sec:internal_comparison}

We first conduct an internal comparison between \spmoe{} 16B and DeepSeek 7B~\citep{deepseek_llm}, a dense language model with 6.9B parameters. 
Ensuring fairness, both models are trained on the same corpus with 2T tokens. 
This enables an accurate assessment of the effectiveness of our MoE architecture, independent of the influence of the training data. 

The evaluation results are presented in Table~\ref{tab:16b_exp_deepseek7b}, yielding the following observations:
(1) On the whole, with about only 40\% of the computations, \spmoe{} 16B achieves comparable performance with DeepSeek 7B. 
(2) \spmoe{} 16B exhibits notable strengths in language modeling and knowledge-intensive tasks such as Pile, HellaSwag, TriviaQA, and NaturalQuestions. 
Given that in an MoE model, FFN parameters are much heavier than attention parameters, these outcomes align with the proposition that FFNs in Transformers exhibit the capability for knowledge memorization~\citep{kn}. 
(3) Compared with the excellent performance on other tasks, \spmoe{} exhibits limitations in addressing multiple-choice tasks. 
This inadequacy stems from the limited attention parameters in \spmoe{} 16B (\spmoe{} 16B has only about 0.5B attention parameters, while DeepSeek 7B has 2.5B attention parameters). 
Our earlier investigation on DeepSeek 7B reveals a positive correlation between the attention capacity and performance on multiple-choice tasks. 
For example, DeepSeek 7B MQA, which is equipped with the multi-query attention mechanism~\citep{mqa}, also struggled in MMLU-like tasks. 
In addition, for a more comprehensive understanding of the training process of \spmoe{} 16B, we also provide the benchmark curves of \spmoe{} 16B and DeepSeek 7B (Dense) during training in Appendix~\ref{sec:app_benchmark_curve} for reference.

Critically, due to the modest number of parameters in \spmoe{} 16B, it enables single-device deployment on a GPU with 40GB of memory. 
With appropriate operator optimizations, it can achieve nearly 2.5 times the inference speed of a 7B dense model.

\begin{table}[ht]
\centering
\setlength{\tabcolsep}{4pt}
\begin{tabular}{@{}l c| c c@{}}
\toprule
\textbf{Metric} & \textbf{\# Shot} & \textbf{LLaMA2 7B} & \textbf{\spmoe{} 16B} \\
\midrule
\# Total Params & N/A & 6.7B & 16.4B \\
\# Activated Params & N/A & 6.7B & 2.8B \\
FLOPs per 4K Tokens & N/A & 187.9T & 74.4T \\
\# Training Tokens & N/A & 2T & 2T \\
\midrule
Pile~(BPB) & N/A & 0.76 & \textbf{0.74} \\
\midrule
HellaSwag~(Acc.) & 0-shot & 75.6 & \textbf{77.1} \\
PIQA~(Acc.) & 0-shot & 78.0 & \textbf{80.2} \\
ARC-easy~(Acc.) & 0-shot & \textbf{69.1} & 68.1 \\
ARC-challenge~(Acc.) & 0-shot & 49.0 & \textbf{49.8} \\
\midrule
RACE-middle~(Acc.) & 5-shot & 60.7 & \textbf{61.9} \\
RACE-high~(Acc.) & 5-shot & 45.8 & \textbf{46.4} \\
DROP~(EM) & 1-shot & \textbf{34.0} & 32.9 \\
\midrule
GSM8K~(EM) & 8-shot & 15.5 & \textbf{18.8} \\
MATH~(EM) & 4-shot & 2.6 & \textbf{4.3} \\
\midrule
HumanEval~(Pass@1) & 0-shot & 14.6 & \textbf{26.8} \\
MBPP~(Pass@1) & 3-shot & 21.8 & \textbf{39.2} \\
\midrule
TriviaQA~(EM) & 5-shot & 63.8 & \textbf{64.8} \\
NaturalQuestions~(EM) & 5-shot & \textbf{25.5} & \textbf{25.5} \\
\midrule
MMLU~(Acc.) & 5-shot & \textbf{45.8} & 45.0 \\
\midrule
WinoGrande~(Acc.) & 0-shot & 69.6 & \textbf{70.2} \\
\midrule
\midrule
CLUEWSC~(EM) & 5-shot & 64.0 & \textbf{72.1} \\
CEval~(Acc.) & 5-shot & 33.9 & \textbf{40.6} \\
CMMLU~(Acc.) & 5-shot & 32.6 & \textbf{42.5} \\
CHID~(Acc.) & 0-shot & 37.9 & \textbf{89.4} \\
\bottomrule
\end{tabular}
\caption{
Comparison between LLaMA2 7B and \spmoe{} 16B. 
With only 39.6\% of computations, \spmoe{} 16B outperforms LLaMA2 7B on the majority of benchmarks. 
}
\label{tab:16b_exp_llama2}
\end{table}

\subsubsection{Comparison with Open Source Models}

\paragraph{Internal Comparison with LLaMA2 7B.}
In the realm of open source models, we mainly compare \spmoe{} 16B with LLaMA2 7B~\citep{llama2}, a well-known and strong open source language model with 6.7B parameters. 
Both \spmoe{} 16B and LLaMA2 7B are pretrained on 2T tokens. 
Compared with LLaMA2 7B, \spmoe{} has 245\% of total parameters but only needs 39.6\% of computations.
The results on our internal benchmarks are presented in Table~\ref{tab:16b_exp_llama2}, leading to the following observations.
(1) Among the evaluated benchmarks, with only about 40\% of computations, \spmoe{} 16B outperforms LLaMA2 7B on the majority of benchmarks.
(2) The math reasoning and code generation capabilities of \spmoe{} 16B are stronger than LLaMA2 7B, attributed to the enriched presence of mathematical and code-related text in our pretraining corpus.
(3) Given the presence of Chinese texts in our pretraining corpus, \spmoe{} 16B exhibits a substantial performance advantage over LLaMA2 7B on Chinese benchmarks.
(4) Despite being trained on fewer English texts, \spmoe{} 16B achieves comparable or better performance compared with LLaMA2 7B on English understanding or knowledge-intensive benchmarks, which demonstrates the exceptional capabilities of \spmoe{} 16B.

\paragraph{Evaluation on Open LLM Leaderboard.}
Beyond our internal evaluations, we also evaluate \spmoe{} 16B on the Open LLM Leaderboard and compare it with other open source models. 
In addition to LLaMA2 7B, we take a broader set of open source models into consideration, including LLaMA 7B~\citep{llama}, Falcon 7B~\citep{falcon}, GPT-J 6B~\citep{gptj}, RedPajama-INCITE 7B and 3B~\citep{redpajama}, Open LLaMA 7B and 3B~\citep{openllama}, OPT 2.7B~\citep{opt}, Pythia 2.8B~\citep{pythia}, GPT-neo 2.7B~\citep{gpt_neo}, and BLOOM 3B~\citep{bloom}. 
The evaluation results, as presented in Figure~\ref{fig:openllm}, show that \spmoe{} 16B consistently outperforms models with similar activated parameters by a large margin. 
Moreover, it achieves comparable performance with LLaMA2 7B, which has approximately 2.5 times the activated parameters. 

\section{Alignment for \spmoe{} 16B}

Previous research indicates that MoE models typically do not emerge significant gains from fine-tuning~\citep{switch,moe_ft_bad}. 
However, \citet{flan_moe} present findings suggesting that MoE models can indeed benefit from instruction tuning.
In order to assess whether \spmoe{} 16B can benefit from fine-tuning, we conduct supervised fine-tuning to construct a chat model based on \spmoe{} 16B. 
The experimental results reveal that \spmoe{} Chat 16B also achieves comparable performance with LLaMA2 SFT 7B and DeepSeek Chat 7B.

\subsection{Experimental Setup}

\paragraph{Training Data.}
For training the chat model, we conduct supervised fine-tuning~(SFT) on our in-house curated data, comprising 1.4M training examples. 
This dataset spans a broad range of categories including math, code, writing, question answering, reasoning, summarization, and more. 
The majority of our SFT training data is in English and Chinese, rendering the chat model versatile and applicable in bilingual scenarios.

\paragraph{Hyper-Parameters.}
During supervised fine-tuning, we set the batch size to 1024 examples and conduct training over 8 epochs using the AdamW optimizer~\citep{adamw}. 
We employ a maximum sequence length of 4K, and pack the training examples as densely as possible until reaching the sequence length limit. 
We do not use dropout for supervised fine-tuning, and simply set a constant learning rate of $10^{-5}$ without incorporating any learning rate scheduling strategy.

\paragraph{Evaluation Benchmarks.}
For the evaluation of the chat models, we employ benchmarks similar to those used in Section~\ref{sec:16b_benchmark}, with the following adjustments:
(1) We exclude Pile~\citep{pile} since chat models are seldom employed for pure language modeling.
(2) We exclude CHID~\citep{chid} due to the observed instability of results, hindering the derivation of solid conclusions.
(3) We additionally include BBH~\citep{bbh} to provide a more comprehensive assessment of the reasoning ability of the chat models.

\begin{table}[ht]
\centering
\setlength{\tabcolsep}{10pt}
\begin{tabular}{@{}l c| c c c@{}}
\toprule
\multirow{2}{*}{\tabincell{c}{\textbf{Metric}}} & \multirow{2}{*}{\tabincell{c}{\textbf{\# Shot}}} & \multirow{2}{*}{\tabincell{c}{\textbf{LLaMA2} \\ \textbf{SFT 7B}}} & \multirow{2}{*}{\tabincell{c}{\textbf{DeepSeek} \\ \textbf{Chat 7B}}} & \multirow{2}{*}{\tabincell{c}{\textbf{\spmoe{}} \\ \textbf{Chat 16B}}} \\
 & & & & \\
\midrule
\# Total Params & N/A & 6.7B & 6.9B & 16.4B \\
\# Activated Params & N/A & 6.7B & 6.9B & 2.8B \\
FLOPs per 4K Tokens & N/A & 187.9T & 183.5T & 74.4T \\
\midrule
HellaSwag~(Acc.) & 0-shot & 67.9 & 71.0 & \textbf{72.2} \\
PIQA~(Acc.) & 0-shot & 76.9 & 78.4 & \textbf{79.7} \\
ARC-easy~(Acc.) & 0-shot & 69.7 & \textbf{70.2} & \textbf{69.9} \\
ARC-challenge~(Acc.) & 0-shot & \textbf{50.8} & 50.2 & 50.0 \\
BBH~(EM) & 3-shot & 39.3 & \textbf{43.1} & 42.2 \\
\midrule
RACE-middle~(Acc.) & 5-shot & 63.9 & \textbf{66.1} & 64.8 \\
RACE-high~(Acc.) & 5-shot & 49.6 & \textbf{50.8} & \textbf{50.6} \\
DROP~(EM) & 1-shot & 40.0 & \textbf{41.7} & 33.8 \\
\midrule
GSM8K~(EM) & 0-shot & \textbf{63.4} & 62.6 & 62.2 \\
MATH~(EM) & 4-shot & 13.5 & 14.7 & \textbf{15.2} \\
\midrule
HumanEval~(Pass@1) & 0-shot & 35.4 & 45.1 & \textbf{45.7} \\
MBPP~(Pass@1) & 3-shot & 27.8 & 39.0 & \textbf{46.2} \\
\midrule
TriviaQA~(EM) & 5-shot & 60.1 & 59.5 & \textbf{63.3} \\
NaturalQuestions~(EM) & 0-shot & \textbf{35.2} & 32.7 & \textbf{35.1} \\
\midrule
MMLU~(Acc.) & 0-shot & \textbf{50.0} & 49.7 & 47.2 \\
\midrule
WinoGrande~(Acc.) & 0-shot & 65.1 & 68.4 & \textbf{69.0} \\
\midrule
\midrule
CLUEWSC~(EM) & 5-shot & 48.4 & 66.2 & \textbf{68.2} \\
CEval~(Acc.) & 0-shot & 35.1 & \textbf{44.7} & 40.0 \\
CMMLU~(Acc.) & 0-shot & 36.9 & \textbf{51.2} & 49.3 \\
\bottomrule
\end{tabular}
\caption{
Comparison among LLaMA2 SFT 7B, DeepSeek Chat 7B and \spmoe{} Chat 16B, with all of these three models fine-tuned on the same SFT data. 
Compared with both 7B dense models, \spmoe{} Chat 16B still achieves comparable or better performance on the majority of benchmarks with only 40\% of computations.
}
\label{tab:16b_exp_chat}
\end{table}

\subsection{Evaluations}

\paragraph{Baselines.}
In order to validate the potential of \spmoe{} 16B after alignment, we conduct supervised fine-tuning for LLaMA2 7B, DeepSeek 7B, and \spmoe{} 16B, where we utilize totally the same fine-tuning data to ensure fairness. 
Correspondingly, we construct three chat models, including LLaMA2 SFT 7B\footnote{We use LLaMA2 SFT to distinguish from the official LLaMA2 Chat~\citep{llama2} model.}, DeepSeek Chat 7B, and \spmoe{} Chat 16B.
Subsequently, we compare \spmoe{} Chat 16B with the other two dense chat models (with about 2.5 times the FLOPs) across a wide range of downstream tasks.

\paragraph{Results.}
The evaluation results are presented in Table~\ref{tab:16b_exp_chat}. 
Our key observations include:
(1) \spmoe{} Chat 16B, while consuming nearly 40\% of computations, achieves comparable performance with 7B dense models across language understanding and reasoning (PIQA, ARC, BBH), machine reading comprehension (RACE), mathematical (GSM8K, MATH), and knowledge-intensive tasks (TriviaQA, NaturalQuestions).
(2) On code generation tasks, \spmoe{} Chat 16B significantly outperforms LLaMA2 SFT 7B, demonstrating notable improvements on HumanEval and MBPP. 
In addition, it also surpasses DeepSeek Chat 7B.
(3) On multiple-choice question answering benchmarks including MMLU, CEval, and CMMLU, \spmoe{} Chat 16B still falls behind DeepSeek Chat 7B, consistent with the observations for the base model~(Section~\ref{sec:internal_comparison}).
However, it is worth noting that, after supervised fine-tuning, the performance gap between \spmoe{} 16B and DeepSeek 7B is narrowed. 
(4) Benefiting from the pretraining on a bilingual corpus, \spmoe{} Chat 16B notably outperforms LLaMA2 SFT 7B on all Chinese benchmarks. 
These results demonstrate the balanced capabilities of \spmoe{} 16B in both Chinese and English, enhancing its versatility and applicability in diverse scenarios.
In conclusion, the evaluation for the chat models highlights the potential of \spmoe{} 16B in benefiting from alignment, and validates its consistent advantages in achieving comparable performance with dense models while using only about 40\% of computations. 

\section{\spmoe{} 145B Ongoing}

Encouraged by the outstanding performance of \spmoe{} 16B, we further undertake a preliminary endeavor to scale up \spmoe{} to 145B. 
In this initial study, \spmoe{} 145B is trained on 245B tokens, but it has demonstrated consistent advantages over the GShard architecture and shown promise to match or exceed the performance of DeepSeek 67B (Dense). 
Furthermore, upon the completion of the final version and full training of \spmoe{} 145B, we also plan to make it publicly available.

\subsection{Experimental Setup}

\paragraph{Training Data and Tokenization.}

For \spmoe{} 145B, we employ exactly the same training corpus and tokenizer as \spmoe{} 16B, with the only difference being that \spmoe{} 145B is trained on 245B tokens for an initial study.

\paragraph{Model Settings.}
For \spmoe{} 145B, we set the number of Transformer layers to 62 and the hidden dimension to 4096. 
We employ the multi-head attention mechanism with a total of 32 attention heads, where each head has a dimension of 128. 
As for initialization, all learnable parameters are randomly initialized with a standard deviation of 0.006.
As in \spmoe{} 16B, we also substitute all FFNs except for the first layer with MoE layers. 
Each MoE layer consists of 4 shared experts and 128 routed experts, where each expert is 0.125 times the size of a standard FFN. 
Each token will be routed to these 4 shared experts and 12 out of 128 routed experts. 
Under this configuration, \spmoe{} 145 has approximately 144.6B total parameters, with the number of activated parameters around 22.2B.

\paragraph{Training Settings.}
We employ the AdamW optimizer~\citep{adamw} with hyper-parameters set to $\beta_1=0.9$, $\beta_2=0.95$, and $\mathrm{weight\_decay}=0.1$. 
For the preliminary study of \spmoe{} 145B, we employ a warmup-and-constant learning rate scheduler. 
Initially, the learning rate linearly increases from 0 to the maximum value during the first 2K steps. 
Subsequently, the learning rate keeps constant during the remaining training process. 
The maximum learning rate for \spmoe{} 145B is set to $3.0 \times 10^{-4}$, and the gradient clipping norm is set to 1.0.
The batch size is set to 4.5K, and with a maximum sequence length of 4K, each training batch contains 18M tokens. 
We train \spmoe{} 145B for 13,000 steps, achieving 245B training tokens. 
Also, we do not use dropout during training. 
We leverage pipeline parallelism to deploy different layers of a model on different devices, and for each layer, all the routed experts will be uniformly deployed on 4 devices (i.e., expert parallelism combined with data parallelism). 
Since we employ expert parallelism for \spmoe{} 145B, the device-level load balance should be considered to reduce the computational bottleneck. 
In response, we set the device-level balance factor to 0.05 to encourage balanced computation across devices. 
Also, we still set a small expert-level balance factor of 0.003 to prevent routing collapse. 

\paragraph{Evaluation Benchmarks.}
We evaluate \spmoe{} 145B on exactly the same internal benchmarks as used for \spmoe{} 16B~(see Section~\ref{sec:16b_benchmark}). 

\begin{table}[!ht]
\centering
\setlength{\tabcolsep}{3pt}
\begin{tabular}{@{}l c| c c c | c@{}}
\toprule
\multirow{2}{*}{\tabincell{c}{\textbf{Metric}}} & \multirow{2}{*}{\tabincell{c}{\textbf{\# Shot}}} & \multirow{2}{*}{\tabincell{c}{\textbf{DeepSeek} \\ \textbf{67B (Dense)}}} & \multirow{2}{*}{\tabincell{c}{\textbf{GShard} \\ \textbf{137B}}} & \multirow{2}{*}{\tabincell{c}{\textbf{\spmoe{}} \\ \textbf{145B}}} & \multirow{2}{*}{\tabincell{c}{\textbf{\spmoe{} 142B} \\ \textbf{(Half Activated)}}} \\
 & & & & \\
\midrule
\# Total Params & N/A & 67.4B & 136.5B & 144.6B & 142.3B \\
\# Activated Params & N/A & 67.4B & 21.6B & 22.2B & 12.2B \\
Relative Expert Size & N/A & N/A & 1 & 0.125 & 0.125 \\
\# Experts & N/A & N/A & 0 + 16 & 4 + 128 & 2 + 128 \\
\# Activated Experts & N/A & N/A & 0 + 2 & 4 + 12 & 2 + 6 \\
FLOPs per 4K Tokens & N/A & 2057.5T & 572.7T & 585.6T & 374.6T \\
\# Training Tokens & N/A & 245B & 245B & 245B & 245B \\
\midrule
Pile~(Loss.) & N/A & 1.905 & 1.961 & \textbf{1.876} & 1.888 \\
\midrule
HellaSwag~(Acc.) & 0-shot & 74.8 & 72.0 & \textbf{75.8} & 74.9 \\
PIQA~(Acc.) & 0-shot & 79.8 & 77.6 & \textbf{80.7} & 80.2 \\
ARC-easy~(Acc.) & 0-shot & 69.0 & 64.0 & \textbf{69.7} & 67.9 \\
ARC-challenge~(Acc.) & 0-shot & \textbf{50.4} & 45.8 & 48.8 & 49.0 \\
\midrule
RACE-middle~(Acc.) & 5-shot & \textbf{63.2} & 59.2 & 62.1 & 59.5 \\
RACE-high~(Acc.) & 5-shot & \textbf{46.9} & 43.5 & 45.5 & 42.6 \\
DROP~(EM) & 1-shot & \textbf{27.5} & 21.6 & \textbf{27.8} & 28.9 \\
GSM8K~(EM) & 8-shot & \textbf{11.8} & 6.4 & \textbf{12.2} & 13.8 \\
MATH~(EM) & 4-shot & 2.1 & 1.6 & \textbf{3.1} & 2.8 \\
\midrule
HumanEval~(Pass@1) & 0-shot & \textbf{23.8} & 17.7 & 19.5 & 23.2 \\
MBPP~(Pass@1) & 3-shot & \textbf{33.6} & 27.6 & \textbf{33.2} & 32.0 \\
\midrule
TriviaQA~(EM) & 5-shot & 57.2 & 52.5 & \textbf{61.1} & 59.8 \\
NaturalQuestions~(EM) & 5-shot & 22.6 & 19.0 & \textbf{25.0} & 23.5 \\
\midrule
MMLU~(Acc.) & 5-shot & \textbf{45.1} & 26.3 & 39.4 & 37.5 \\
\midrule
WinoGrande~(Acc.) & 0-shot & 70.7 & 67.6 & \textbf{71.9} & 70.8 \\
\midrule
\midrule
CLUEWSC~(EM) & 5-shot & 69.1 & 65.7 & \textbf{71.9} & 72.6 \\
CEval~(Acc.) & 5-shot & \textbf{40.3} & 26.2 & 37.1 & 32.8 \\
CMMLU~(Acc.) & 5-shot & \textbf{40.6} & 25.4 & 35.9 & 31.9 \\
CHID~(Acc.) & 0-shot & 88.5 & 86.9 & \textbf{90.3} & 88.3 \\
\bottomrule
\end{tabular}
\caption{
Comparison among DeepSeek 67B (Dense) and MoE models at the scale of about 140B total parameters. 
In the lines of ``\# Experts'' and ``\# Activated Experts'', $a$ + $b$ denotes $a$ shared experts and $b$ routed experts, respectively. 
\textbf{Bold} font indicates the best or near the best performance excluding the last column.
\spmoe{} 145B, and even \spmoe{} 142B (Half Activated) that has only a half of activated expert parameters, outperform GShard 137B by a large margin. 
Moreover, with 28.5\% of computations, \spmoe{} 145B achieves comparable performance with DeepSeek 67B. 
}
\label{tab:145b_exp_base}
\end{table}

\subsection{Evaluations}

\paragraph{Baselines.}

Apart from \textbf{\spmoe{} 145B}, we consider three additional models for comparison. 
\textbf{DeepSeek 67B (Dense)} is a dense model with 67.4B total parameters (refer to \citet{deepseek_llm} for the model and training details). 
\textbf{GShard 137B} shares the same hidden dimension and number of layers as \spmoe{} 145B, but follows the GShard architecture. 
Note that \spmoe{} 145B aligns the intermediate hidden dimension in each expert to a multiple of 64 for computation efficiency, so its model size is 6\% larger than GShard 137B.
\textbf{\spmoe{} 142B (Half Activated)} has a similar architecture to \spmoe{} 145B, but it contains only 2 shared experts, and only 6 out of 128 routed experts are activated. 
It is noteworthy that all compared models, including \spmoe{} 145B, share the same training corpus. 
In addition, all MoE models in the comparison are trained from scratch and share the same training hyper-parameters.

\paragraph{Results.}
From the evaluation results presented in Table~\ref{tab:145b_exp_base}, we have the following observations:
(1) 
Despite having comparable total parameters and computations, \spmoe{} 145B significantly outperforms GShard 137B, highlighting the advantages of the \spmoe{} architecture again.
(2) 
On the whole, with only 28.5\% of computations, \spmoe{} 145B achieves comparable performance with DeepSeek 67B (Dense). 
Consistent with the findings from \spmoe{} 16B, \spmoe{} 145B exhibits remarkable strengths in language modeling and knowledge-intensive tasks, but with limitations in multiple-choice tasks.
(3) 
At a larger scale, the performance of \spmoe{} 142B (Half Activated) does not lag behind too much from \spmoe{} 145B. 
In addition, despite having only a half of activated expert parameters, \spmoe{} 142B (Half Activated) still match the performance of DeepSeek 67B (Dense), with only 18.2\% of computations. 
It also outperforms GShard 137B, which aligns with the conclusion from Section~\ref{sec:validation_analysis_specialization}. 

\section{Related Work}

The Mixture of Experts~(MoE) technique is first proposed by \citet{ori_moe1,ori_moe2} to deal with different samples with independent expert modules. 
\citet{moe} introduce MoE into language model training and build a large-scale LSTM-based~\citep{lstm} MoE models. 
As Transformer become the most popular architecture for NLP, many attempts extend FFNs in a Transformer as MoE layers to build MoE language models. 
GShard~\citep{gshard} and Switch Transformer~\citep{switch} are pioneers which employ learnable top-2 or top-1 routing strategies to scale the MoE language models to an extremely large scale.
Hash Layer~\citep{hash} and StableMoE~\citep{stablemoe} use fixed routing strategies for more stable routing and training. 
\citet{ec} propose an expert-choice routing strategy, where each token can be assigned to different numbers of experts. 
\citet{st_moe} focus on the issues of training instability and fine-tuning difficulty in MoE models, and propose ST-MoE to overcome these challenges. 
In addition to research on MoE architectures and training strategies, recent years have also witnessed the emergence of numerous large-scale language or multimodal models~\citep{m6,glam,pangu_sigma,openmoe} based on existing MoE architectures. 
By and large, most of the previous MoE models are based on conventional top-1 or top-2 routing strategies, leaving large room for improving expert specialization. 
In response, our \spmoe{} architecture aims to improve the expert specialization to the utmost extent.

\section{Conclusion}

In this paper, we introduce the \spmoe{} architecture for MoE language models, with the objective of achieving ultimate expert specialization. 
Through fine-grained expert segmentation and shared expert isolation, \spmoe{} achieves significantly higher expert specialization and performance compared with prevailing MoE architectures. 
Starting with a modest scale of 2B parameters, we validate the advantages of \spmoe{}, demonstrating its capability to approach the upper bound performance for MoE models. 
Furthermore, we provide empirical evidence to show that \spmoe{} has a higher level of expert specialization than GShard. 

Scaling up to a larger scale of 16B total parameters, we train \spmoe{} 16B on 2T tokens and demonstrate its outstanding performance comparable with DeepSeek 7B and LLaMA2 7B, with only about 40\% of computations.
Additionally, supervised fine-tuning is conducted for alignment to construct an MoE chat model based on \spmoe{} 16B, further showing its adaptability and versatility.
Further, we perform a preliminary exploration to scale \spmoe{} to 145B parameters. 
We find that \spmoe{} 145B still keeps substantial advantages over the GShard architecture, and demonstrates comparable performance with DeepSeek 67B, using only 28.5\% (maybe even 18.2\%) of computations.

For research purposes, we release the model checkpoint of \spmoe{} 16B to the public, which can be deployed on a single GPU with 40GB of memory. 
We aspire for this work to provide valuable insights for both academia and industry, and contribute to the accelerated advancement of large-scale language models.

\bibliography{main}

\newpage
\appendix

\section*{Appendices}

\section{Overview of Hyper-Parameters}
\label{sec:app_hyper_params}

We present the overview of hyper-parameters for \spmoe{} across various sizes in Table~\ref{tab:hyper_params}.

\begin{table}[ht]
\centering
\footnotesize
\setlength{\tabcolsep}{2pt}
\begin{tabular}{@{}c c c c c c c c c c@{}}
\toprule
\multirow{2}{*}{\# Params} & \multirow{2}{*}{\# Layers} & \multirow{2}{*}{\tabincell{c}{Hidden \\ Size}} & \multirow{2}{*}{\tabincell{c}{\# Attn \\ Heads}} & \multirow{2}{*}{\tabincell{c}{\# Shared \\ Experts}} & \multirow{2}{*}{\tabincell{c}{\# Routed \\ Experts}} & \multirow{2}{*}{\tabincell{c}{Relative \\ Expert Size}} & \multirow{2}{*}{\tabincell{c}{Sequence \\ Length}} & \multirow{2}{*}{\tabincell{c}{Batch Size \\ (Sequence)}} & \multirow{2}{*}{\tabincell{c}{Learning \\ Rate}} \\
 & & & & & & & & \\
\midrule
~~~~2.0B  & 9 & 1280 & 10 & 1 & 63 (7 activated) & 0.25 & 2048 & 2048 & 1.08e-3 \\
~~16.4B & 28 & 2048 & 16 & 2 & 64 (6 activated) & 0.25 & 4096 & 4608 & ~~4.2e-4 \\
144.6B & 62 & 4096 & 32 & 4 & 128 (12 activated) & 0.125 & 4096 & 4608 & ~~3.0e-4 \\
\bottomrule
\end{tabular}
\caption{
Overview of hyper-parameters for \spmoe{} across various sizes.
The relative expert size is in comparison to a standard FFN.
}
\label{tab:hyper_params}
\end{table}

\section{Comparing \spmoe{} with Larger Models}
\label{sec:app_compare_larger}

Comparisons among \spmoe{}, GShard$\times 1.2$, and GShard$\times 1.5$ are shown in Table~\ref{tab:valid_exp_large_gshard}. 
Comparisons among \spmoe{}, Dense$\times 4$, and Dense$\times 16$ are shown in Table~\ref{tab:valid_exp_large_dense}. 

\begin{table}[ht]
\centering
\setlength{\tabcolsep}{8pt}
\begin{tabular}{@{}l c| c c c c@{}}
\toprule
\textbf{Metric} & \textbf{\# Shot} & \textbf{GShard$\times 1.2$} & \textbf{GShard$\times 1.5$} & \textbf{\spmoe{}} \\
\midrule
Relative Expert Size & N/A & 1.2 & 1.5 & 0.25 \\
\# Experts & N/A & 0 + 16 & 0 + 16 & 1 + 63 \\
\# Activated Experts & N/A & 0 + 2 & 0 + 2 & 1 + 7 \\
\# Total Expert Params & N/A & 2.3B & 2.8B & 1.9B \\
\# Activated Expert Params & N/A & 0.28B & 0.35B & 0.24B \\
\# Training Tokens & N/A & 100B & 100B & 100B \\
\midrule
Pile~(Loss) & N/A & 1.824 & \textbf{1.808} & \textbf{1.808} \\
\midrule
HellaSwag~(Acc.) & 0-shot & 53.7 & 54.4 & \textbf{54.8} \\
PIQA~(Acc.) & 0-shot & 71.8 & 71.1 & \textbf{72.3} \\
ARC-easy~(Acc.) & 0-shot & 46.8 & 47.3 & \textbf{49.4} \\
ARC-challenge~(Acc.) & 0-shot & 31.7 & \textbf{34.1} & \textbf{34.3} \\
\midrule
RACE-middle~(Acc.) & 5-shot & 43.7 & \textbf{46.4} & 44.0 \\
RACE-high~(Acc.) & 5-shot & 31.9 & \textbf{32.4} & 31.7 \\
\midrule
HumanEval~(Pass@1) & 0-shot & 3.7 & 3.0 & \textbf{4.9} \\
MBPP~(Pass@1) & 3-shot & 2.4 & \textbf{2.6} & 2.2 \\
\midrule
TriviaQA~(EM) & 5-shot & 15.2 & 15.7 & \textbf{16.6} \\
NaturalQuestions~(EM) & 5-shot & 4.5 & 4.7 & \textbf{5.7} \\
\bottomrule
\end{tabular}
\caption{
\centering Comparison between \spmoe{} and larger GShard models. 
}
\label{tab:valid_exp_large_gshard}
\end{table}

\begin{table}[ht]
\centering
\setlength{\tabcolsep}{8pt}
\begin{tabular}{@{}l c| c c c c@{}}
\toprule
\textbf{Metric} & \textbf{\# Shot} & \textbf{Dense$\times 4$} & \textbf{Dense$\times 16$} & \textbf{\spmoe{}} \\
\midrule
Relative Expert Size & N/A & 1 & 1 & 0.25 \\
\# Experts & N/A & 4 + 0 & 16 + 0 & 1 + 63 \\
\# Activated Experts & N/A & 4 + 0 & 16 + 0 & 1 + 7 \\
\# Total Expert Params & N/A & 0.47B & 1.89B & 1.89B \\
\# Activated Expert Params & N/A & 0.47B & 1.89B & 0.24B \\
\# Training Tokens & N/A & 100B & 100B & 100B \\
\midrule
Pile~(Loss) & N/A & 1.908 & \textbf{1.806} & \textbf{1.808} \\
\midrule
HellaSwag~(Acc.) & 0-shot & 47.6 & \textbf{55.1} & \textbf{54.8} \\
PIQA~(Acc.) & 0-shot & 70.0 & 71.9 & \textbf{72.3} \\
ARC-easy~(Acc.) & 0-shot & 43.9 & \textbf{51.9} & 49.4 \\
ARC-challenge~(Acc.) & 0-shot & 30.5 & 33.8 & \textbf{34.3} \\
\midrule
RACE-middle~(Acc.) & 5-shot & 42.4 & \textbf{46.3} & 44.0 \\
RACE-high~(Acc.) & 5-shot & 30.7 & \textbf{33.0} & 31.7 \\
\midrule
HumanEval~(Pass@1) & 0-shot & 1.8 & 4.3 & \textbf{4.9} \\
MBPP~(Pass@1) & 3-shot & 0.2 & \textbf{2.2} & \textbf{2.2} \\
\midrule
TriviaQA~(EM) & 5-shot & 9.9 & \textbf{16.5} & \textbf{16.6} \\
NaturalQuestions~(EM) & 5-shot & 3.0 & \textbf{6.3} & 5.7 \\
\bottomrule
\end{tabular}
\caption{
\centering Comparison between \spmoe{} and larger dense baselines. 
}
\label{tab:valid_exp_large_dense}
\end{table}

At a larger scale of 13B total parameters, we also compare \spmoe{} with GShard$\times 1.2$ and GShard$\times 1.5$, and show results in Table~\ref{tab:valid_exp_large_gshard_13b}. 
At a larger scale, \spmoe{} even outperforms GShard$\times 1.5$ distinctly.

\begin{table}[!ht]
\centering
\setlength{\tabcolsep}{8pt}
\begin{tabular}{@{}l c| c c c c@{}}
\toprule
\textbf{Metric} & \textbf{\# Shot} & \textbf{GShard$\times 1.2$} & \textbf{GShard$\times 1.5$} & \textbf{\spmoe{}} \\
\midrule
Relative Expert Size & N/A & 1.2 & 1.5 & 0.25 \\
\# Experts & N/A & 0 + 16 & 0 + 16 & 1 + 63 \\
\# Activated Experts & N/A & 0 + 2 & 0 + 2 & 1 + 7 \\
\# Total Expert Params & N/A & 15.9B & 19.8B & 13.3B \\
\# Activated Expert Params & N/A & 2.37B & 2.82B & 2.05B \\
\# Training Tokens & N/A & 100B & 100B & 100B \\
\midrule
HellaSwag~(Acc.) & 0-shot & 66.6 & 67.7 & \textbf{69.1} \\
PIQA~(Acc.) & 0-shot & 75.6 & \textbf{76.0} & \textbf{75.7} \\
ARC-easy~(Acc.) & 0-shot & 56.8 & 56.8 & \textbf{58.8} \\
ARC-challenge~(Acc.) & 0-shot & \textbf{39.9} & 37.6 & 38.5 \\
\midrule
RACE-middle~(Acc.) & 5-shot & 51.6 & 50.6 & \textbf{52.4} \\
RACE-high~(Acc.) & 5-shot & 37.4 & 36.3 & \textbf{38.5} \\
\midrule
HumanEval~(Pass@1) & 0-shot & 6.1 & 6.1 & \textbf{9.8} \\
MBPP~(Pass@1) & 3-shot & 7.0 & \textbf{11.6} & 10.6 \\
\midrule
TriviaQA~(EM) & 5-shot & 36.5 & 36.7 & \textbf{38.2} \\
NaturalQuestions~(EM) & 5-shot & 12.6 & 12.1 & \textbf{13.7} \\
\bottomrule
\end{tabular}
\caption{
\centering Comparison between \spmoe{} and larger GShard models at a larger scale. 
}
\label{tab:valid_exp_large_gshard_13b}
\end{table} 

\section{Training Benchmark Curves of \spmoe{} 16B}
\label{sec:app_benchmark_curve}

\begin{figure}[ht]
\centering
\includegraphics[width=0.99\linewidth]{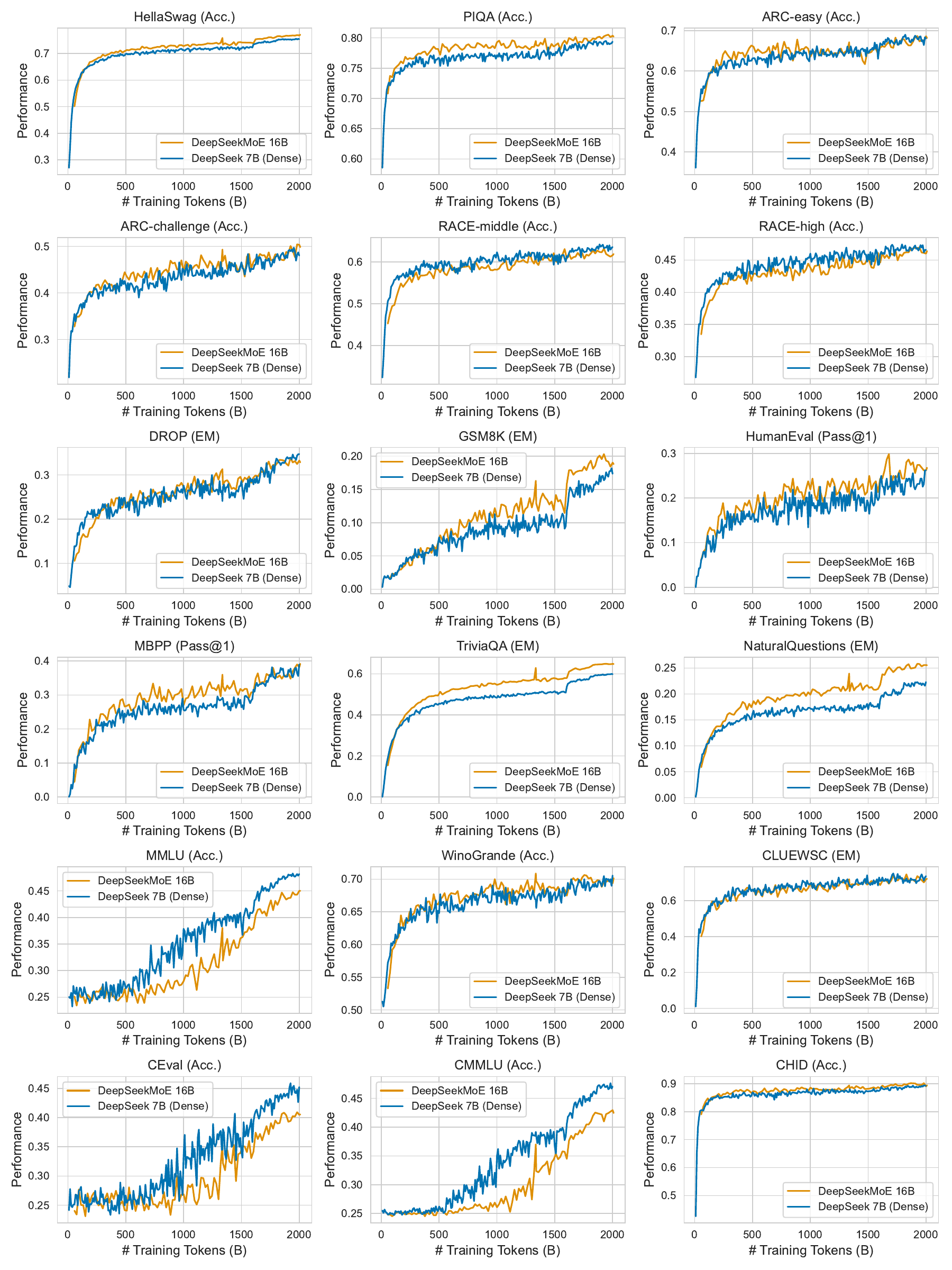}
\caption{
\centering Benchmark curves during training of \spmoe{} 16B and DeepSeek 7B (Dense). 
}
\label{fig:benchmark_curve}
\end{figure}

We present the benchmark curves during training of \spmoe{} 16B and DeepSeek 7B (Dense) in Figure~\ref{fig:benchmark_curve} for reference.

\end{CJK*}
\end{document}